\documentclass[sigconf]{acmart}
\usepackage{colortbl}

\usepackage{multirow}
\usepackage{array} 
\usepackage{tabularx}
\usepackage{makecell} 

\usepackage{enumitem}
\usepackage{hanging}

\usepackage{algorithm}
\usepackage{algpseudocode}

\usepackage{tcolorbox}
\usepackage{fancyvrb}
\usepackage{listings}

\usepackage{subfigure}
\usepackage{tikz} 

\usepackage{lipsum}

\definecolor{lightblue}{rgb}{0.21, 0.49, 0.74}

\AtBeginDocument{%
  }

\copyrightyear{2026}
\acmYear{2026}
\setcopyright{acmlicensed}
\setcctype{by}
\acmConference[MM '26] {Proceedings of the 35th ACM International Conference on Multimedia}{November 10--14, 2026}{Rio de Janeiro, Brazil.}
\acmBooktitle{Proceedings of the 34th ACM International Conference on Multimedia (MM '26), November 10--14, 2026, Rio de Janeiro, Brazil}
\acmISBN{979-8-4007-2213-4/2026/11}
\acmDOI{10.1145/3767308.3835261}

\begin{document}

\title[A Visual-Feedback Framework for Structured SVG Generation in Complex Document and Meeting Scenarios]{GVR-Coder: A Visual-Feedback Framework for Structured SVG Generation in Complex Document and Meeting Scenarios}


\author{Yiming Xu}
\orcid{0009-0003-0918-4037}
\email{xym30@mail.ustc.edu.cn}
\affiliation{%
 \institution{University of Science and Technology of China}
   \city{Hefei}
 \state{Anhui}
 \country{China}
}

\author{Jihua Kang}
\orcid{0009-0000-3843-1402}
\email{kangjihua@bytedance.com}
\affiliation{%
 \institution{ByteDance Inc.}
   \city{Shanghai}
 \country{China}
}

\author{Chunsai Du}
\orcid{0009-0000-6994-7241}
\email{duchunsai@bytedance.com}
\affiliation{%
 \institution{ByteDance Inc.}
  \city{Shanghai}
 \country{China}
}

\author{Qifan Zhang}
\orcid{0009-0006-1101-3781}
\email{work.qfzhang@gmail.com}
\affiliation{%
 \institution{ByteDance Inc.}
 \city{Shanghai}
 \country{China}
}

\author{Wangqiu Zhou}
\orcid{0000-0002-2915-4324}
\email{rafazwq@hfut.edu.cn}
\affiliation{%
 \institution{Hefei University of Technology}
 \city{Hefei}
 \state{Anhui}
 \country{China}
}

\author{Yiting Wu}
\orcid{0000-0002-8517-4100}
\email{wuyitingde@gmail.com}
\affiliation{%
 \institution{ByteDance Inc.}
 \city{Shanghai}
 \country{China}
}

\author{Tianqi Li}
\orcid{0009-0007-6996-6291}
\email{tianqi.li@bytedance.com}
\affiliation{%
 \institution{ByteDance Inc.}
 \city{Shanghai}
 \country{China}
}

\author{Qi Song}
\orcid{0000-0002-1726-7858}
\authornote{Corresponding author.}
\email{qisong09@ustc.edu.cn}
\affiliation{%
 \institution{University of Science and Technology of China}
 \city{Hefei}
 \state{Anhui}
 \country{China}
}

\renewcommand{\shortauthors}{Xu et al.}

\begin{abstract}
In demanding professional environments and meeting review scenarios, lengthy text often imposes a high cognitive load. To facilitate efficient information communication, transforming verbose text into logically clear diagrams is essential. Scalable Vector Graphics (SVG) provide an effective representation for this purpose due to their editability and resolution independence.
However, current research on Text-to-SVG generation remains hindered by three major challenges: (1) the scarcity of datasets for complex, logic-rich diagrams; (2) the absence of explicit layout priors, which leads to chaotic spatial arrangements; and (3) the lack of fine-grained visual feedback to validate rendered outputs and correct aesthetic defects.
To address these challenges, at the data level, we introduce DocMeetSVG-100K, a large-scale SVG dataset tailored for document authoring and meeting review scenarios. At the model level, we propose GVR-Coder, a novel framework designed to generate high-quality logical diagrams from lengthy professional texts. 
Specifically, we adopt a curriculum-driven rejection sampling fine-tuning to progressively enhance the model’s capability in modeling complex structures, while explicitly incorporating layout constraint knowledge during training. In addition, we introduce reinforcement learning from dual rendering feedback, a mechanism that provides implicit feedback through reward signals to jointly optimize structural complexity and visual aesthetics. Furthermore, we design a generate-verify-repair agent loop, which improves generation quality through explicit, fine-grained feedback and targeted refinement.
Extensive experiments demonstrate that GVR-Coder outperforms competitive baselines and reliably produces logically coherent and visually appealing diagrams. Code and data are available at https://github.com/CurryaNa/GVR-Coder.
\end{abstract}

\ccsdesc[500]{Computing methodologies~Computer vision}

\keywords{Large Language Models; Scalable Vector Graphics; Generative Models; Reinforcement Learning}


\maketitle

\section{Introduction}

\begin{figure}[h]
  \centering
  \includegraphics[width=1.0\linewidth]{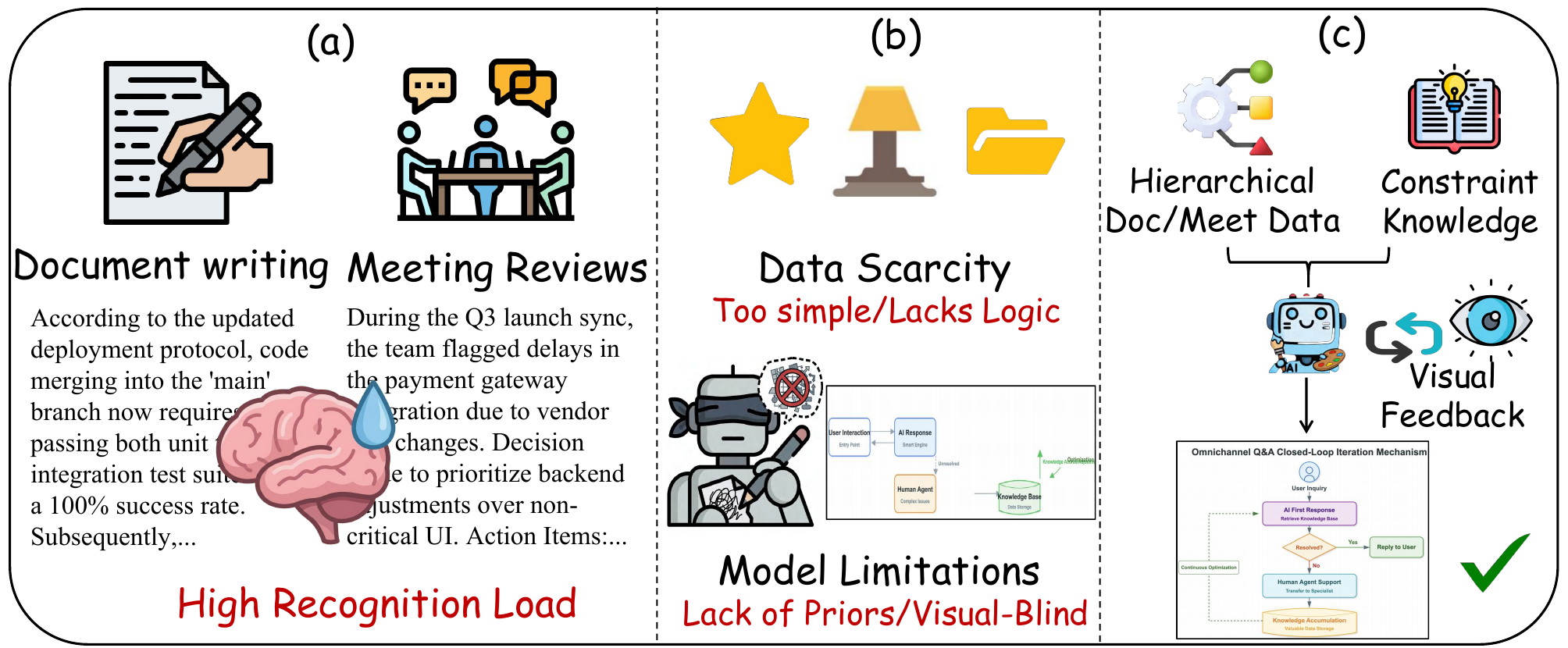}
  \caption{Motivation. (a) Verbose texts impose high cognitive loads. (b) Existing methods struggle with overly simplistic outputs, lack of layout priors, and visual blindness. (c) Our framework resolves this via a curated hierarchical dataset, explicit knowledge, and iterative visual feedback.}
  \Description{A woman and a girl in white dresses sit in an open car.}
  \label{example}
\end{figure}

In modern professional office environments, such as document writing~\cite{zhu2026paperbanana, zhu2026autofigure, li2026ragtrack} and meeting reviews~\cite{wang2025research,asthana2025summaries,li2026cadtrack}, effective information organization and presentation are essential for distilling and communicating complex content. However, lengthy textual documents and dense meeting records often impose considerable cognitive load (Figure \ref{example}(a)), making core logical structures and key insights difficult to identify efficiently. In contrast, structured visual representations—such as diagrams, mind maps, and statistical charts—provide a more intuitive means of conveying procedural logic, task decomposition, and data relationships, thereby improving communication efficiency~\cite{zhao2025vincicoder,xu2025lgc}.
Due to its resolution independence and editability, 
Scalable Vector Graphics (SVG) has become an important technical medium for such visual expressions~\cite{li2025unisvg,chen2025svgenius}. 

While SVGs offer distinct advantages, traditional design workflows~\cite{tian2025review} rely on professional tools and manual editing, which are labor-intensive and poorly suited for rapid iteration. Therefore, enabling the automatic generation of logically coherent and layout-consistent SVG diagrams from verbose texts remains a critical open research problem. Recently, the proliferation of Artificial Intelligence Generated Content (AIGC) has driven research into automated Text-to-SVG generation, primarily diverging into two paradigms. Optimization-based approaches (e.g., \cite{thamizharasan2024nivel,xing2023diffsketcher,zhang2024text,xing2024svgfusion}) typically generate SVGs by iteratively optimizing differentiable pixel renderers. However, these methods suffer from low efficiency and often produce disorganized SVG that lack semantic editability. In contrast, recent LLM-based approaches \cite{xing2025empowering,hazimeh2026semantic,wu2023iconshop,wang2025svgen,liu2024hrvda,yang2025omnisvg,wang2025robosvg, wu2025chat2svg} reframe SVG generation as a code synthesis task. By fine-tuning large (visual) language models and introducing specialized tokens to capture SVG structures, these methods significantly boost generation efficiency and code editability, thereby emerging as the 
mainstream direction in Text-to-SVG research.

Although LLM-based methods have achieved significant progress, they still face three major challenges. 
 \textbf{First, the scarcity of high-quality Text-to-SVG datasets specifically curated for complex, high-logic scenarios}. Current research predominantly focuses on simple, isolated icons (Figure \ref{example} (b)), which fails to support the generation of sophisticated diagrams required in scenarios like meeting reviews and document authoring.
 \textbf{Second, the absence of a "designing mind" for constraint-based knowledge guidance.} While current models excel at generating valid code, they inherently lack explicit aesthetic and layout knowledge priors. Consequently, the generated SVGs often suffer from chaotic spatial arrangements.
\textbf{Finally, the lack of a "perceptive eye" for fine-grained visual feedback.} Existing methods primarily treat SVG generation as a pure code synthesis task but never 
validate the rendered outputs visually (Figure \ref{example} (b)), which prevents models from perceiving aesthetic defects and performing repairs.
To this end, we propose an innovative solution that augments both the data and model levels (Figure \ref{example}(c)).  
\textbf{For Challenge 1}, we construct DocMeetSVG-100K, a large-scale, finely annotated Text-to-SVG dataset with progressive difficulty levels, filling the gap in document authoring and meeting scenarios. 
\textbf{For Challenge 2}, we introduce a curriculum-driven rejection sampling fine-tuning strategy and incorporate layout knowledge. This design provides "lessons" of varying difficulty to progressively cultivate the model’s understanding of SVG diagrams, along with structured priors to provide rigorous layout constraints, thereby averting chaotic arrangements.
\textbf{For Challenge 3},  we establish a perception system that combines implicit and explicit feedback. During training,  we develop a reinforcement learning strategy with dual rendering rewards, which internalizes structural complexity and visual aesthetics into the model’s generation intuition via reward signals. During inference,  we propose a generate-verify-repair agent loop. The verifier evaluates the rendered SVG to provide fine-grained attribution and suggestions, identifying not just the quality gap but also the underlying causes ('why') and specific errors ('where'). This directs the repair model to perform targeted refinement for high-quality outputs.
To sum up, our contributions are:

\textbf{(1) DocMeetSVG-100K Dataset:} To alleviate the cognitive load of long-text processing in office scenarios, we construct DocMeetSVG 100K, a large-scale office Text-to-SVG dataset specifically curated for logic diagram design, along with a automated evaluation protocal that combines aesthetic standards and factual consistency.

\textbf{(2) GVR-Coder SVG Generation Framework:}  We propose GVR-Coder, a novel framework that synergizes progressive knowledge internalizing and iterative visual feedback. By integrating knowledge-enhanced curriculum learning, reinforcement learning with dual-reward signals, and a multi-agent repair loop, GVR-Coder bridges the gap between code synthesis and visual perception.

\textbf{(3) Three-Stage Optimization Strategy:}
First, we introduce a curriculum-driven training strategy that incorporates layout knowledge to progressively enhance the model's grasp of logic-rich structural priors; second, we design a novel hybrid reward function for reinforcement learning, providing signals for both structural complexity and visual quality optimization; finally, we develop a generate–verify–repair agent loop to iteratively provide fine-grained visual feedback and refine the outputs.

\textbf{(4) Experimental Results:} Experimental results demonstrate that GVR-Coder not only outperforms traditional approaches but also surpasses larger-scale LLMs both qualitatively and quantitatively. 
Further ablation studies provide strong 
empirical evidence for the effectiveness of the core components. The source code and partial data are provided in the supplementary material.

\begin{figure*}[htbp]
    \centering
\includegraphics[width=1.0\textwidth]{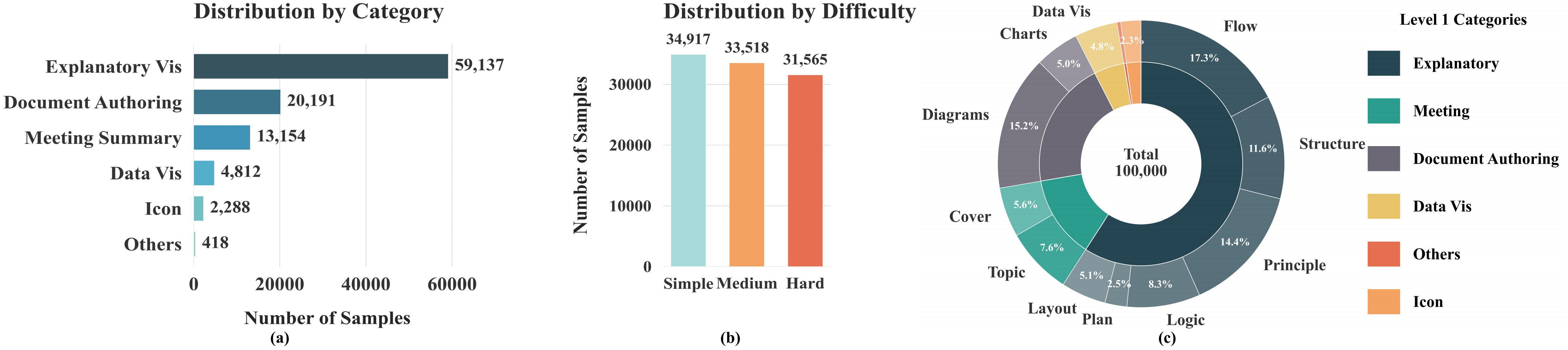}
    \caption{Statistics of our dataset DocMeetSVG-100K. (a) shows the sample count across six primary categories. (b) presents the volume of samples categorized by difficulty levels. (c) illustrates the proportional breakdown of sub-categories.}
    \label{fig:data_dis}
\end{figure*}

\section{Related Work}

\subsection{Vector Graphics Generation}

SVG generation methods can be broadly divided into optimization-based and LLM-based paradigms. Optimization-based methods use differentiable rasterizers to optimize SVG parameters, such as CLIP-guided sketch~\cite{radford2021learning} generation in CLIPDraw~\cite{frans2022clipdraw} and CLIPasso~\cite{vinker2022clipasso}, or diffusion-driven iterative refinement in VectorFusion~\cite{jain2023vectorfusion} and SVGDreamer~\cite{xing2024svgdreamer}. However, these methods often produce entangled paths with limited semantic editability, while their iterative optimization also reduces efficiency. Recently, LLM-based methods~\cite{leon2025gpt, he2026vfig, gemini_30, liu2024deepseek, team2026kimi, zeng2026glm, wang2025internsvg, rodriguez2023starvector} have become increasingly attractive due to their semantic understanding and code generation ability. LLM4SVG~\cite{xing2024empowering} and OmniSVG~\cite{yang2025omnisvg} improve controllability through semantic tokens or structural-geometric decoupling, while ReasonSVG~\cite{xing2025reason} and SVGen~\cite{wang2025svgen} further introduce reasoning or reward-based optimization. Despite these advances, existing approaches remain limited, struggling either with outputs that lack structural diversity and complex logical organization, or with a fundamental absence of SVG layout priors and visual feedback.

\subsection{SVG Datasets}
While early works such as FIGR-8-SVG \cite{clouatre2019figr} and SVG-Icons8 \cite{carlier2020deepsvg} provide massive collections of basic icons, they are typically restricted to monochrome, single-object designs lacking compositional structure. To enhance visual expressiveness, datasets like ColorSVG-100K \cite{chen2024svgbuilder}, LLM4SVG \cite{xing2025empowering}, and MMSVG-2M~\cite{yang2025omnisvg} introduced colorful vector graphics and intricate anime designs, leveraging multimodal models to construct text prompts. Recently, SVG-Stack \cite{rodriguez2023starvector}, SVG-Sophia \cite{wang2026reliable}, and SVG-1M~\cite{wang2025svgen} have further scaled text-SVG alignment to the million level, incorporating semantic annotations and Chain-of-Thought (CoT) enhancement.
However, while these datasets \cite{kocetkov2022stack,wang2023deepvecfont} have advanced the field in terms of scale and artistic visual richness, they are fundamentally oriented toward general UI assets. They lack topological relationships (e.g., flowcharts), and strict hierarchical nesting (e.g., mind maps) required in professional office settings.

\section{DocMeetSVG-100K Dataset Construction}
\label{sec:dataset}
To bridge the structural and logical gaps in existing SVG datasets, we introduce DocMeetSVG-100K, a large-scale dataset specifically designed for professional document and meeting scenarios. 
To ensure authenticity and diversity, we collect captions from real-world sources, including meeting records, professional office documents, and the Arena conversation dataset. Each input is formulated as a structured tuple of raw text, application scenario, and target diagram type. Since different diagram categories require different structural and layout patterns, we adopt type-specific construction pipelines rather than a single unified prompt template. For each caption, Gemini-3-Pro generates two independent SVG candidates, which are evaluated by a Judge Model from visual, layout, topology, and semantic-consistency perspectives. Imperfect samples are further refined by a specialized Repair Model until they meet the predefined quality standards. After generation and repair, we remove duplicates and conduct stratified human verification, where 30\% of samples in each difficulty level are manually audited for visual quality, semantic faithfulness, and layout correctness. For evaluation, the 320 test cases are collected from real documents and meeting records outside the training set, and all test samples are fully manually validated before benchmarking. As shown in Figure~\ref{fig:data_dis}, DocMeetSVG-100K contains 100,000 Text-SVG pairs across six primary categories with balanced difficulty levels, supporting logic-rich SVG diagram generation.

\begin{figure*}[!thbp]
    \centering
    \includegraphics[width=1.0\textwidth]{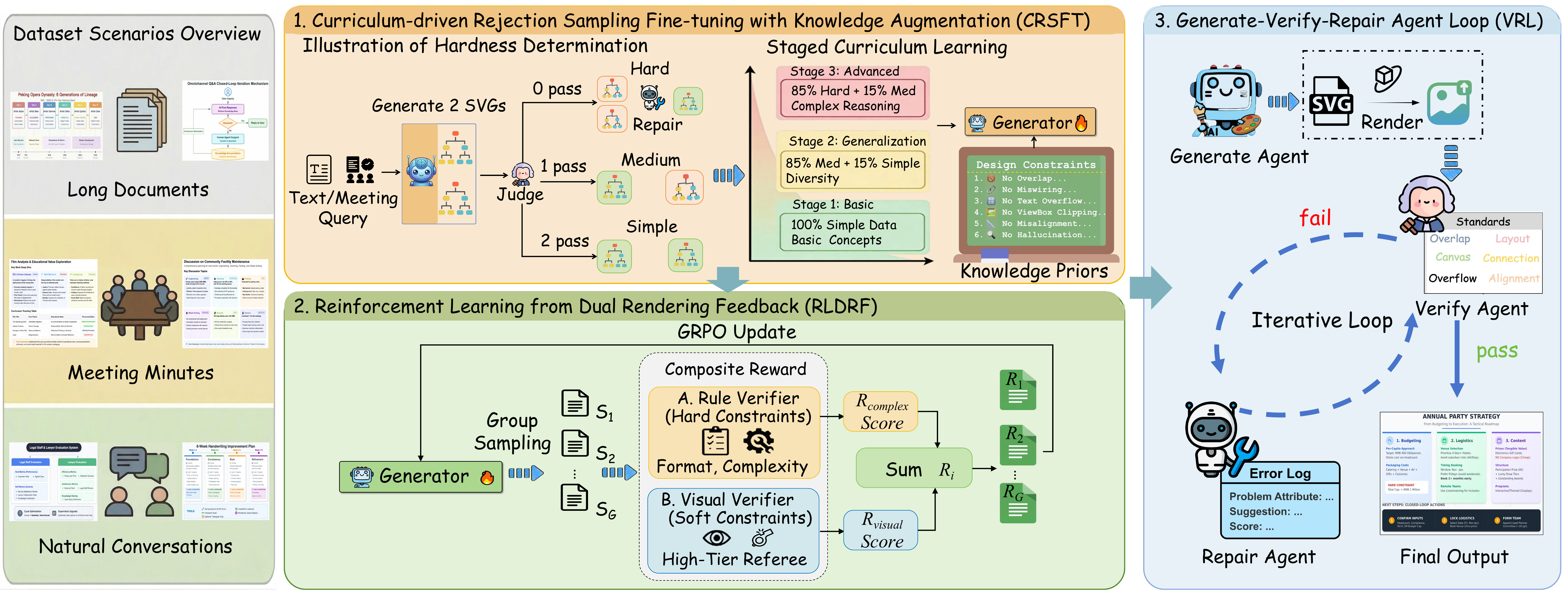}
    \caption{The overall framework: Dataset scenarios and GVR-Coder workflow.
}
    \label{fig:double_column}
\end{figure*}

\section{Method}
In modern professional scenarios, such as document authoring and post-meeting reviews, users are often overwhelmed by lengthy and unstructured texts. The goal of our task is to automatically transform these verbose inputs into logically clear and visually intuitive SVG diagrams, thereby organizing procedural logic, and explicitly presenting core themes. 
To achieve this, we propose GVR-Coder, a visual-feedback framework tailored for complex SVG generation. As illustrated in Figure ~\ref{fig:double_column}, our framework integrates three core stages to achieve logically rigorous and aesthetically superior outputs.

\subsection{Curriculum-driven Rejection Sampling Fine-tuning with Knowledge Augmentation}
As described in Sec. \ref{sec:dataset}, our basic idea is to harness the powerful generation capabilities of LLMs to assist in constructing a high-quality training dataset. 
However, while advanced models like Gemini-3-Pro demonstrate remarkable potential in code synthesis, their success in generating logically rigorous SVGs is often stochastic.

Therefore, to transform this generative diversity into high-quality supervision signals, we decide to employ rejection sampling to improve the quality of the fine-tuning data. 
Specifically, for each input text caption, we first prompt the model to generate a set of candidate responses. 
Then, an automated verifier filters these candidates, rejecting those with defects and retaining only the perfect outputs to construct a high-confidence Supervised Fine-Tuning (SFT) dataset. 
Moreover, directly training the model on a uniformly mixed dataset can easily lead to convergence failure. 
To this end, we introduce a curriculum learning strategy to progressively enhance the model’s learning capability, preventing it from being overwhelmed by overly difficult data at the initial stage.


Combining these insights, we propose Curriculum-driven Rejection Sampling Fine-Tuning (CRSFT). In our implementation, we deliberately set the number of responses to exactly two to balance generative diversity with computational overhead. Conveniently, the rejection sampling process naturally stratifies the data into three difficulty levels based on the sampling success ratio: 
(1) \textbf{Simple}. Both samples are perfect, covering fundamental SVG structures and attributes.
(2) \textbf{Medium}. Only one sample is perfect, introducing generative diversity.
(3) \textbf{Hard}. Both samples exhibit flaws, typically corresponding to complex tasks or multi-step reasoning. These instances are optimized using a specialized \textit{Repair Model}.

Following the cognitive principle of ``learning from simple to complex'', we implement a three-stage curriculum learning. 
Stage 1 (Basic) utilizes 100\% simple data to build a foundational understanding of basic SVG syntax and design. Stage 2 (Generalization) introduces a mix of 85\% medium and 15\% simple data to enhance generative diversity and cross-scenario generalization. Finally, Stage 3 (Advanced) incorporates 85\% hard and 15\% medium data to strengthen the model's generative capability for complex logical tasks.

While current models excel at generating valid code, their lack of layout priors often leads to chaotic arrangements. To compensate for this inherent deficiency,
we introduce a constraint-aware knowledge augmentation module. Explicit design constraints—including priors such as collision avoidance (via bounding box constraints), connectivity logic (via anchor-point alignment), and factual fidelity (via anti-hallucination rules)—are injected into the prompt to guide the model's spatial reasoning and minimize structural defects.

\subsection{Reinforcement Learning from Dual Rendering Feedback}
While CRSFT improves basic generation ability, token-level likelihood optimization alone cannot ensure visual fidelity. To address this, we propose Reinforcement Learning from Dual Rendering Feedback (RLDRF), which employs GRPO to optimize a hybrid reward balancing aesthetics and structural complexity. The aesthetic reward improves visual quality but may encourage overly simplified outputs to avoid layout errors. Therefore, we introduce a complexity reward to prevent such degeneration by maintaining element density comparable to the ground truth. This balance enables the model to generate diagrams that are both visually appealing and structurally rich.
The reward components are defined as follows.

\paragraph{(1) Aesthetic Reward:} For each generated SVG, we first perform a validity check. Non-renderable code is penalized with a score of 0. Valid images are evaluated by a high-tier VLM (Gemini-3-pro) across six dimensions: element overlap, connection issues, text overflow, layout clutter, alignment styles, and content occlusion. The reward is calculated using a deductive scoring system:
\begin{equation}
    R_{visual} = \frac{\max(0, 6 - N_{errors})}{6},
\end{equation}
where $N_{errors}$ represents the number of detected visual defects.

\paragraph{(2) Complexity Reward:} We introduce a complexity matching reward, which counts structural tags $n$ (including \textit{path, circle, line, etc.}) in both generated and reference code. A threshold of 0.8 is applied to ensure the generated result matches the target complexity:
\begin{equation}
R_{complex} = 
\begin{cases} 
1.0 & \text{if } n_{gen} \geq 0.8 \cdot n_{ref} \\
\frac{n_{gen}}{0.8 \cdot n_{ref}} & \text{otherwise} 
\end{cases}.
\end{equation}

This approach ensures the generated SVG maintains a density of elements comparable to the ground truth.

The total reward  $R$ combines the aforementioned rewards:
\begin{equation}
R_{total} = R_{visual} + R_{complex}.
\end{equation}

Through GRPO optimization on 8,000 hard samples, the model is capable of generating diagrams that meet aesthetic expectations while maintaining structural richness.

\subsection{Generate-Verify-Repair Agent Loop}

While the RLDRF stage effectively aligns the model's global generation policy with visual preferences, it fundamentally relies on scalar reward signals. Consequently, the model learns whether a generated SVG possesses high overall quality, but lacks explicit attribution regarding where or why specific local defects occur, leaving it with no opportunity to repair its single-pass outputs. To address this limitation during inference, we introduce a multi-agent loop.  This framework simulates a pedagogical process of ``student submission $\rightarrow$ teacher feedback $\rightarrow$ student revision $\rightarrow$ teacher score''. Unlike the implicit score-based guidance in training, this iterative framework utilizes a verifier to provide fine-grained, attribute-level visual feedback, offering actionable suggestions that guide a specialized repair agent to perform targeted refinements. The core components and iterative logic are as follows:

Generate Agent:  
Powered by the model optimized in the preceding CRSFT and RLDRF stages, this agent serves as the starting point to receive a constraint-refined instruction (Caption) and synthesize an initial SVG code $P_0 = \text{Agent}_{\text{gen}}(\text{Caption})$. To bridge code and visual perception, the generated program is subsequently rendered into a visual image $I_t = \text{Agent}_{\text{render}}(P_t)$ at each $t$-th iteration.


Verify Agent: 
Leveraging the multimodal capabilities of Gemini-3-Pro, this agent analyzes the original caption, the SVG code $P_t$, and the rendered image $I_t$, in order to detect logical deviations or visual artifacts. The evaluation process is formulated as:
\begin{equation}
\{D_t, F_t, S_t\} = \text{VLM}_{\text{verify}}(\text{Caption}, P_t, I_t),
\end{equation}
where $D_t$ denotes the decision result (pass/fail), $F_t$ represents feedback information containing problem attributes and revision suggestions, and $S_t$ denotes the current evaluation score.

Repair Agent:
To equip the model with robust code revision capabilities, this agent is trained on a curated dataset of approximately 20,000 repair instances. This training corpus is constructed from two complementary sources: automated corrections generated by Gemini-3-Pro and manual refinements executed by human annotators, who adjust visual components using the Inkscape software. During the loop, the agent
takes the current SVG $P_t$ and the verifier's feedback $F_t$ as inputs to reconstruct and optimize the SVG accordingly. The process is defined as:
\begin{equation}
P_{t+1} = \text{Agent}_{\text{repair}}(P_t, F_t).
\end{equation}

The loop is executed at most three times, as empirical observations indicate that performance gains plateau beyond this point, with additional iterations offering negligible incremental utility. 

\section{Experiment}
To evaluate the performance of our proposed method for SVG generation tasks, we conduct comprehensive experiments on the constructed DocMeetSVG-100K dataset. Specifically, we reserve 320 samples as an independent test set, while utilizing the remaining data for training. To systematically assess model performance across diverse real-world scenarios, this test set is composed of four distinct benchmarks: \textbf{Arena\_human\_50}, an open-ended multilingual dialogue-driven setting; \textbf{Meeting\_cover\_100}, focusing on structured meeting cover layout generation; \textbf{Meeting\_topic\_50}, targeting abstract topic-to-structure modeling; and \textbf{Svg\_with\_rules\_120}, which imposes strict topological and layout constraints derived from professional documentation and academic papers, representing a highly creative generation scenario.

We provide the complete implementation in our code repository, where the training details (including full-parameter fine-tuning, GRPO training, and the ms-swift \cite{zhao2025swift} framework) are specified.

\subsection{Evaluation Protocols}
\label{sec:evaluation}

\subsubsection{Automated Evaluation System}

(1) Aesthetic and Normative Scoring:
Standard Text-to-Image metrics such as FID and CLIPScore often correlate poorly with the logical and topological requirements of document-level diagrams. We therefore adopt Gemini-3-Pro as VLM-as-Judge to provide nuanced visual feedback.


\begin{figure}[htbp]
    \centering
\includegraphics[width=0.5\textwidth]{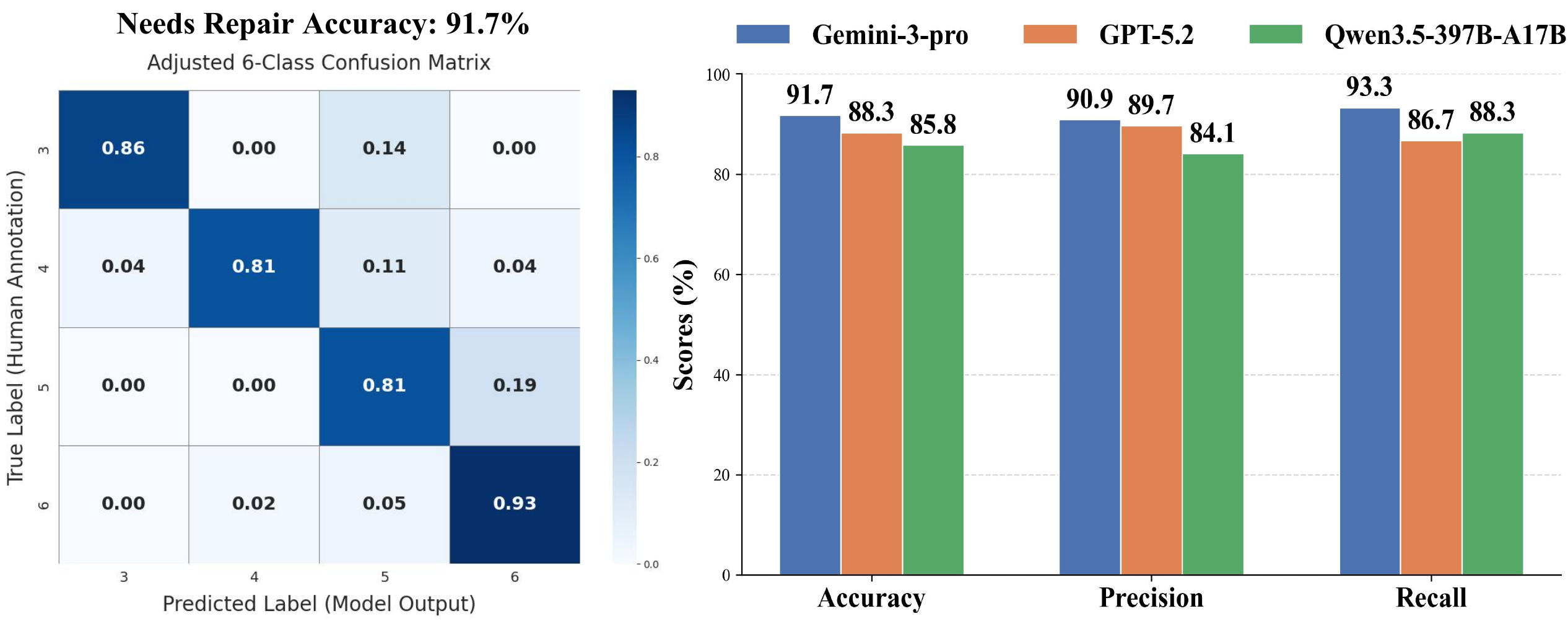}
    \caption{Verifier's Reliability (Left) and Robustness (Right).}
    \label{fig:eval}
\end{figure}
To bridge the gap between automated checking and manual review, we classify visual defects into three dimensions: boundaries and completeness, topology and logic, and layout and aesthetics. These dimensions cover six error types: canvas clipping, content occlusion, text overflow, connector issues, misalignment, and cluttered layout. We then adopt a deduction-based scoring mechanism:
\begin{equation}
Score = \max(0, 6 - N),
\end{equation}
where $N$ represents the number of detected errors.

(2) Factuality Scoring:
The score is assigned to measure semantic alignment across three core dimensions: Zero Hallucination, No Omission, and Logical Consistency.
1 Point: The SVG satisfies all three criteria, faithfully reflecting the caption without redundant content.
0 Point: The image fails in any dimension. For more detailed prompts, please refer to the supplementary materials.


\begin{table*}[t]
\setlength{\tabcolsep}{3pt}
\small
\renewcommand{\arraystretch}{0.95}
\caption{Quantitative comparison with SOTA models. The \colorbox[HTML]{DCDCDC}{gray} and \colorbox[HTML]{D7F6FF}{blue} color rows denote the performances of open-source models and GVR-Coder. Best in \textbf{bold}, runner-up \underline{underlined}. In addition to the \textit{Average score} (As), we emphasize the perfect \textit{Acceptance rate} (Ar), which measures the proportion of fully usable outputs. And scores are categorized by quality: 0-2 (Low), 3-4 (Medium), and 5-6 (High). Furthermore, we introduce \textit{SVG Validity} (first-pass syntactic validity) to assess code-level robustness.}
\label{tab:main_results}
\centering
\resizebox{\textwidth}{!}{
\begin{tabular}{l|ccccc|ccccc|ccccc|ccccc|c}
\toprule
\multirow{2}{*}{Model} & \multicolumn{5}{c|}{\textbf{Arena\_human\_50}} & \multicolumn{5}{c|}{\textbf{Meeting\_cover\_100}} & \multicolumn{5}{c|}{\textbf{Meeting\_topic\_50}} & \multicolumn{5}{c|}{\textbf{Svg\_with\_rules\_120}} & \textbf{SVG} \\
\cmidrule{2-22}
    & As & Ar & Low & Med.  & High & As & Ar & Low & Med.  & High & As & Ar & Low & Med.  & High & As & Ar & Low & Med.  & High & \textbf{Valid.(\%)} \\ 
\midrule
\multicolumn{22}{l}{\textbf{Traditional SVG Models}} \\ 
\midrule
 SVGGen & 80.0 & 16.0 & 0 & 17 & 33 & 72.0  & 0.0 & 0 & 68 & 32 & 78.0 & 0.0 & 0 & 15 & 35 & 76.0 & 0.0 & 0 & 52 & 68 & 89.9\\
 LLM4SVG & 60.0 & 2.0 & 5 & 35 & 8 & 50.0 & 0.0 & 23 & 75 & 2 & 56.0 & 0.0 & 6 & 41 & 3 & 50.0 & 0.0 & 30 & 85 & 5 & 76.5\\
\midrule

\multicolumn{22}{l}{\textbf{Open-Source Models}} \\ 
\midrule
\rowcolor[HTML]{DCDCDC} InternVL-3.5-38B & 70.0 & 4.0 & 1 & 31 & 18 & 61.0  & 2.0 & 3 & 91 & 6 & 66.0 & 0.0 & 0 & 43 & 7 & 64.0 & 3.3 & 1 & 98 & 21 & 88.7\\
\rowcolor[HTML]{DCDCDC} Deepseek-V3.2 & 74.7 & 24.0 & 0 & 30 & 20 & 67.0 & 4.0 & 2 & 74 & 24  & 68.0 & 8.0 & 0 & 41 & 9 & 66.5 & 6.0 & 3 & 90 & 27 & 92.6\\
\rowcolor[HTML]{DCDCDC} Qwen3.5-397B-A17B & 80.3 & 30.0 & 0 & 19 & 31 & 81.7 & 29.0 & 1 & 30 & 69 & 84.0 & 44.0 & 0 & 19 & 31 & 81.3 & 35.0 & 2 & 42 & 76  & 92.8\\
\midrule
\multicolumn{22}{l}{\textbf{Closed-Source Models}} \\
\midrule

GPT-5.1   & 70.3 & 4.0 & 0 & 33 & 17 & 68.8 & 4.0 & 0 & 74 & 26 &71.0 & 8.0 & 0 & 34 & 16 &  68.1 & 5.8 & 2 & 87 & 31 & 95.6\\
Gemini-2.5-pro & 78.0 & 18.0 & 0 & 23 & 27 & 84.3 & 44.0 & 0 & 35 & 65 & 71.3 & 12.0 & 0 & 36 & 14 & 78.9 & 28.0 & 0 & 56 & 64 & 94.4\\
GLM-5 & 79.0 & 28.0 & 0 & 25 & 25 & 84.3 & 38.0 & 0 & 30 & 70 & 83.7 & 34.0 & 0 & 16 & 34 & 83.2 & 34.2 & 0 & 38 & 82 & 93.0 \\
Kimi-k2.5 & 81.7 & 28.0 & 0 & 17 & 33 & 87.5 & 46.0 & 0 & 20 & 80 & 86.0 & 48.0 & 0 & 13 & 37 & 81.4 & 29.2 & 0 & 45 & 75 & 95.9\\
GPT-5.5 & 78.3 & 36.0 & 0 & 27 & 23 & 86.3 & 55.0 & 0 & 34 & 66 & 80.7 & 42.0 & 0 & 25 & 25 & 72.9 & 27.5 & 2 & 77 & 41 & 94.8\\
Gemini-3-pro & 83.3 & 38.0 & 0 & 17 & 33 & 92.8 & 69.0 & 0 & 11 & 89 & 89.0 & 56.0 & 0 & 11 & 39 & 86.0 & 50.0 & 0 & 34 & 86 & 97.2\\
\midrule
\multicolumn{22}{l}{\textbf{GVR-Coder}} \\
\midrule
Qwen3-14B  & 66.0 & 6.0 & 1 & 41 & 8 & 67.5 &  1.0 & 0 & 74 & 26  & 64.7 & 2.0 & 0 & 42 & 8 & 62.1 & 1.0 & 3 & 101 & 16 & 76.3\\
\rowcolor[HTML]{D7F6FF} GVR-Coder-14B-RSFT  & 87.3 & 42.0 & 0 & 11 & 39 & 94.0 &  71.0 & 0 & 9 & 91  & 89.0 & 60.0 & 0 & 12 & 38 & 87.4 & 52.5 & 0 & 32 & 88 & 96.9\\
\rowcolor[HTML]{D7F6FF} GVR-Coder-14B-RLDRF  & 93.3 & 74.0 & 0 & 5 & 45 & 95.0 & 74.0 & 0  & 6 & 94 & 96.5 & 82.0 & 0 & 3 & 47 & 88.0 & 56.6 & 1 & 32 & 87 & \underline{99.4}\\
\rowcolor[HTML]{D7F6FF} GVR-Coder-14B-VRL   &\textbf{96.0} & \textbf{82.0} & 0 & 1 & 49 & \underline{97.7} & \underline{90.0} & 0 & 4 & 96 & \underline{97.0} & \underline{88.0} & 0 & 3 & 47 & \underline{95.3} & \underline{81.6} & 0 &  12 & 108 & N/A \\
\midrule
Qwen3-32B & 70.0 & 6.0 & 1 & 37 & 12 & 73.8 & 13.0 & 0 & 61 & 39 & 70.7 & 10.0 & 0 & 34 & 16 & 66.0 & 3.3 & 4 & 92 & 24 & 89.0\\ 
\rowcolor[HTML]{D7F6FF} GVR-Coder-32B-RSFT & 89.0 & 48.0 & 0 & 7 & 43 &  95.0 & 68.0 & 0 & 6 & 94 & 93.0 & 70.0 & 0 & 7 & 43  & 86.0 & 48.3 & 0 & 36 & 84 & 97.8\\
\rowcolor[HTML]{D7F6FF} GVR-Coder-32B-RLDRF & 94.0 &  \underline{76.0} & 0 & 4 & 46 & 94.8 & 75.0 & 0 & 13 & 87 & \underline{97.0} & 84.0 & 0 & 2 & 48 & 88.0 & 54.2 & 0 & 29 & 91 & \textbf{99.7} \\
\rowcolor[HTML]{D7F6FF} GVR-Coder-32B-VRL & \underline{94.7} & \textbf{82.0} & 0 & 7 & 43 & \textbf{100.0} & \textbf{100.0}  & 0 & 0 & 100 & \textbf{97.3} & \textbf{90.0} & 0 & 3 & 47 & \textbf{96.9} & \textbf{88.3} & 0 & 7 & 113 & N/A 
\\
\bottomrule 
\end{tabular}
}
\end{table*}

\subsubsection{Reliability of VLM-as-a-Judge}
We compare the VLM's performance against human experts on 200 samples (Figure \ref{fig:eval}). The VLM-as-a-judge achieved a 90\% agreement rate with humans in aesthetic and normative "repair vs. no repair" decisions. Fine-grained scoring showed strong alignment, with diagonal proportions in the six-class confusion matrix ranging from 0.81 to 0.93. 
Regarding factual consistency, it achieved an F1 score of 0.87 compared to human annotations.
Meanwhile (Right), various models show consistently high performance, confirming our protocol is model-agnostic and supports cost-effective open-source deployment.

\subsection{Quantitative Analysis}
\begin{figure}[!htbp]
    \centering
    \includegraphics[width=0.8\linewidth]{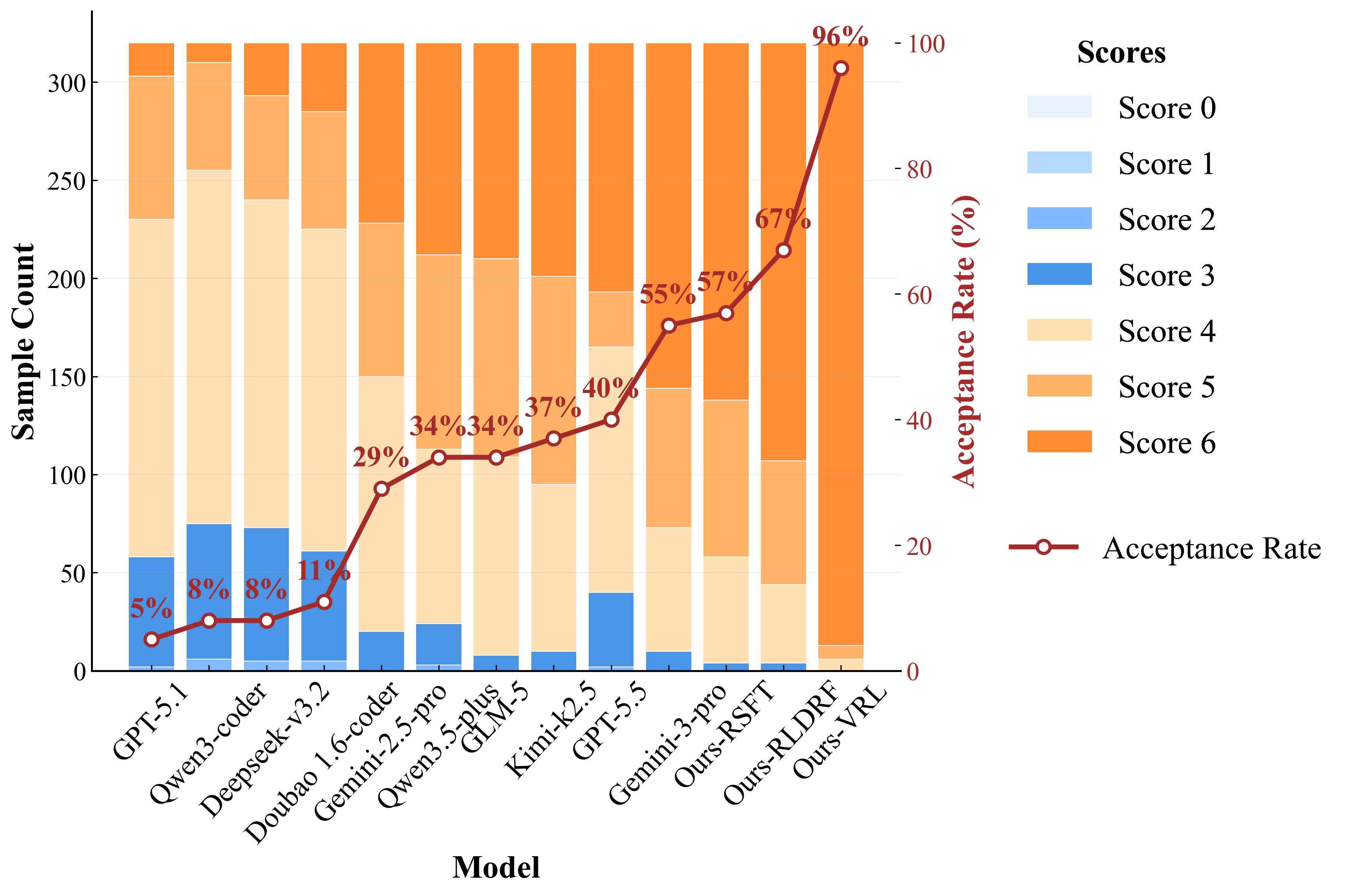}
    \caption{Score distribution and Acceptance Rate on 320 evaluation samples.}
    \label{fig:aesthetic_score}
\end{figure}

As shown in Table \ref{tab:main_results} and Figure \ref{fig:aesthetic_score}, GVR-Coder variants consistently outperform all baselines. Traditional methods fail in document-level tasks (Ar $\approx$ 0), as their icon-centric training data lacks the priors for complex hierarchical organization.

\begin{figure}[!htbp]
    \centering
    \includegraphics[width=0.6\linewidth]{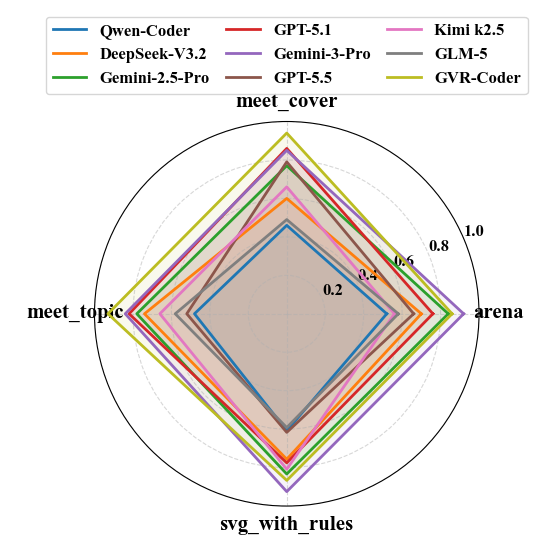}
    \caption{Factuality Performance Radar Map.}
    \label{fig:fact_radar}
\end{figure}

Compared to SOTA closed-source models, GVR-Coder achieves superior accept ratio in open-ended, multilingual dialogues (82.0\% on Arena), strict constraint tasks (88.0\% on Svg\_with\_rules), and meeting scenarios (up to 100\%). These results highlight our model's reliability in topic extraction and hierarchical organization, while effectively mitigating text overflow, element overlap, and connection conflicts. Regarding factual consistency, GVR-Coder ranks top in meeting datasets and second in open-ended settings, where semantic boundaries are less explicit (Figure \ref{fig:fact_radar}).

Performance gains follow a clear stage-wise pattern: (1) RSFT establishes a robust foundation (As > 86.0) through curriculum learning; (2) RLDRF bridges the modality gap using visual rewards, significantly boosting Ar; and (3) VRL resolves specific layout defects via fine-grained visual feedback and targeted repair. Furthermore, GVR-Coder achieves a first-pass rendering success rate > 99\% by enforcing syntactic constraints during RL, substantially reducing regeneration costs in real-world deployment.

\subsection{Cross-Evaluator Consistency and Quality--Latency Trade-off}

Although Gemini-3-Pro is used as the primary VLM-as-Judge, our results are not evaluator-specific. As shown in Table~\ref{tab:cross_evaluator}, re-evaluation with Gemini-3-Pro, GPT-5.2, and Qwen3.5-397B yields stable rankings, with GVR-Coder consistently achieving the highest acceptance rate. This suggests that the improvement is not an artifact of a single Gemini-based scoring protocol.

\begin{figure}[!htbp]
    \centering
    \includegraphics[width=0.95\linewidth]{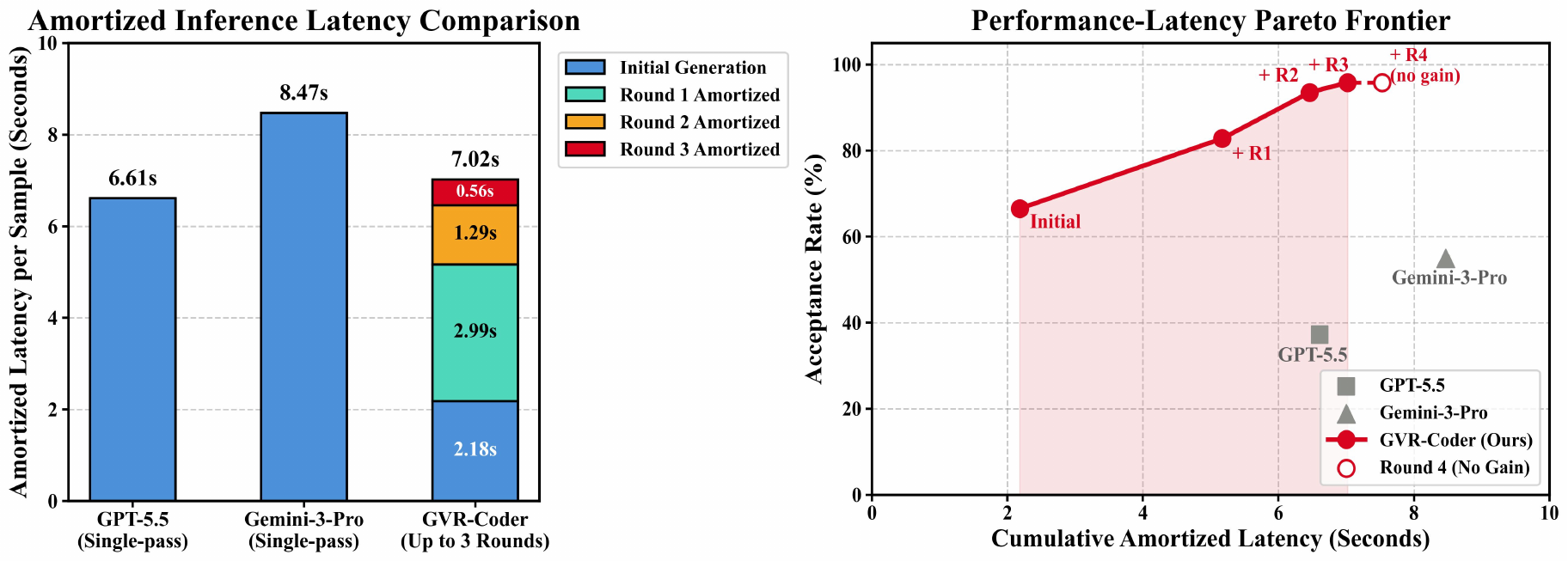}
    \caption{Inference efficiency and performance trade-off.}
    \label{fig:trade_off}
\end{figure}

As shown in Fig.~\ref{fig:trade_off}, GVR-Coder achieves 96\% acceptance with only 7.02s amortized latency, which is lower than Gemini-3-Pro single-pass generation (8.47s) and close to GPT-5.5 (6.61s). The Pareto frontier further shows that the first three repair rounds steadily improve acceptance with limited latency overhead, while the fourth round brings almost no additional gain. This supports our choice of limiting VRL to three iterations and demonstrates a favorable quality--latency trade-off.

\begin{table}[t]
\centering
\caption{Cross-Verifier robustness under different VLMs.}
\label{tab:cross_evaluator}
\small 
\setlength{\tabcolsep}{3.5pt} 
\renewcommand{\arraystretch}{0.85} 
\setlength{\aboverulesep}{0ex}
\setlength{\belowrulesep}{0ex}
\begin{tabular}{lcccccc}
\toprule
\multirow{2}{*}{Model} 
& \multicolumn{2}{c}{Gemini-3-Pro}
& \multicolumn{2}{c}{GPT-5.2}
& \multicolumn{2}{c}{Qwen3.5-397B} \\
\cmidrule(lr){2-3} \cmidrule(lr){4-5} \cmidrule(lr){6-7}
& As & Ar & As & Ar & As & Ar \\
\midrule
GLM-5.2                 & 83.0 & 34\% & 81.8 & 38\% & 82.0 & 35\% \\
GPT-5.1               & 69.1 & 5\%  & 68.7 & 5\%  & 65.2 & 4\%  \\
Gemini-3-Pro          & 88.5 & 55\% & 88.1 & 56\% & 87.4 & 53\% \\
\midrule
\rowcolor[HTML]{D7F6FF} GVR-Coder-14B-RSFT    & 89.0 & 57\% & 90.0 & 58\% & 89.0 & 56\% \\
\rowcolor[HTML]{D7F6FF} GVR-Coder-14B-RLDRF   & \underline{91.1} & \underline{65\%} & \underline{90.7} & \underline{64\%} & \underline{90.0} & \underline{62\%} \\
\rowcolor[HTML]{D7F6FF} GVR-Coder-14B-VRL   & \textbf{98.1} & \textbf{96\%} & \textbf{99.1} & \textbf{97\%} & \textbf{97.6} & \textbf{94\%} \\
\bottomrule
\end{tabular}
\end{table}


\begin{table}[h]
\centering
\renewcommand{\arraystretch}{0.95}
\caption{Ablation study of the GVR-Coder framework.}
\resizebox{0.98\columnwidth}{!}{
\begin{tabular}{lccc}
\toprule
Model & Average Score $\uparrow$ & Accept Ratio $\uparrow$ & Usability Rate $\uparrow$ \\
\midrule
Base Model (Vanilla SFT) & 86.2 & 49.1 & 76.0 \\
+ Curriculum Learning & 88.5 & 54.6 & 78.1 \\
+ Knowledge Injection & 90.2 & 56.8 & 80.3 \\
+ RLDRF & 91.1 & 67.2 & 83.0 \\
\midrule
\textbf{+ VRL (Ours)} & \textbf{97.6} & \textbf{95.8} & \textbf{97.2} \\
\bottomrule
\end{tabular}
}
\label{tab:ablation}
\end{table}

\subsection{Ablation study}

We start from the base model and incrementally add each component to demonstrate their cumulative contributions (Table \ref{tab:ablation}).
We examine these improvements through the lens of two core metrics: Accept Ratio (Score 6) and Usability Rate (Score $\ge 5$).

\textbf{Analysis of Curriculum Learning.}
We graded instruction data difficulty based on rejection sampling success rates. With this "easy-to-hard" strategy, the accept ratio increases from the 49.0\% baseline to 55.0\%. It allows the model to steadily build capabilities from simple primitives to complex topologies. This prevents the model from being overwhelmed by too hard tasks in early stages, significantly enhancing convergence stability and alignment quality.

\textbf{Analysis of Constraint Knowledge Injection.}
To mitigate text models' spatial perception deficiencies, we injected constraint knowledge (e.g., collision avoidance and overflow prevention) during training. Incorporating this prior further boosts the accept ratio from 54.6\% to 56.8\%. This confirms that purely data-driven methods struggle with strict geometric constraints. Converting defects into design rules provides clear decision boundaries.

\textbf{Analysis of RLDRF.} To enhance complex diagram generation, we apply GRPO on 8,000 hard samples. Ablation results show that RLDRF improves the accept ratio from 56.8\% to 67.2\%, highlighting the necessity of implicit visual feedback via reward signals when handling complex scenarios. A more detailed analysis of the reward design and its effects is provided in the Supplementary Material.

\begin{figure}[!htbp]
    \centering
    \hspace*{-0.4cm} 
    \includegraphics[width=0.45\textwidth]{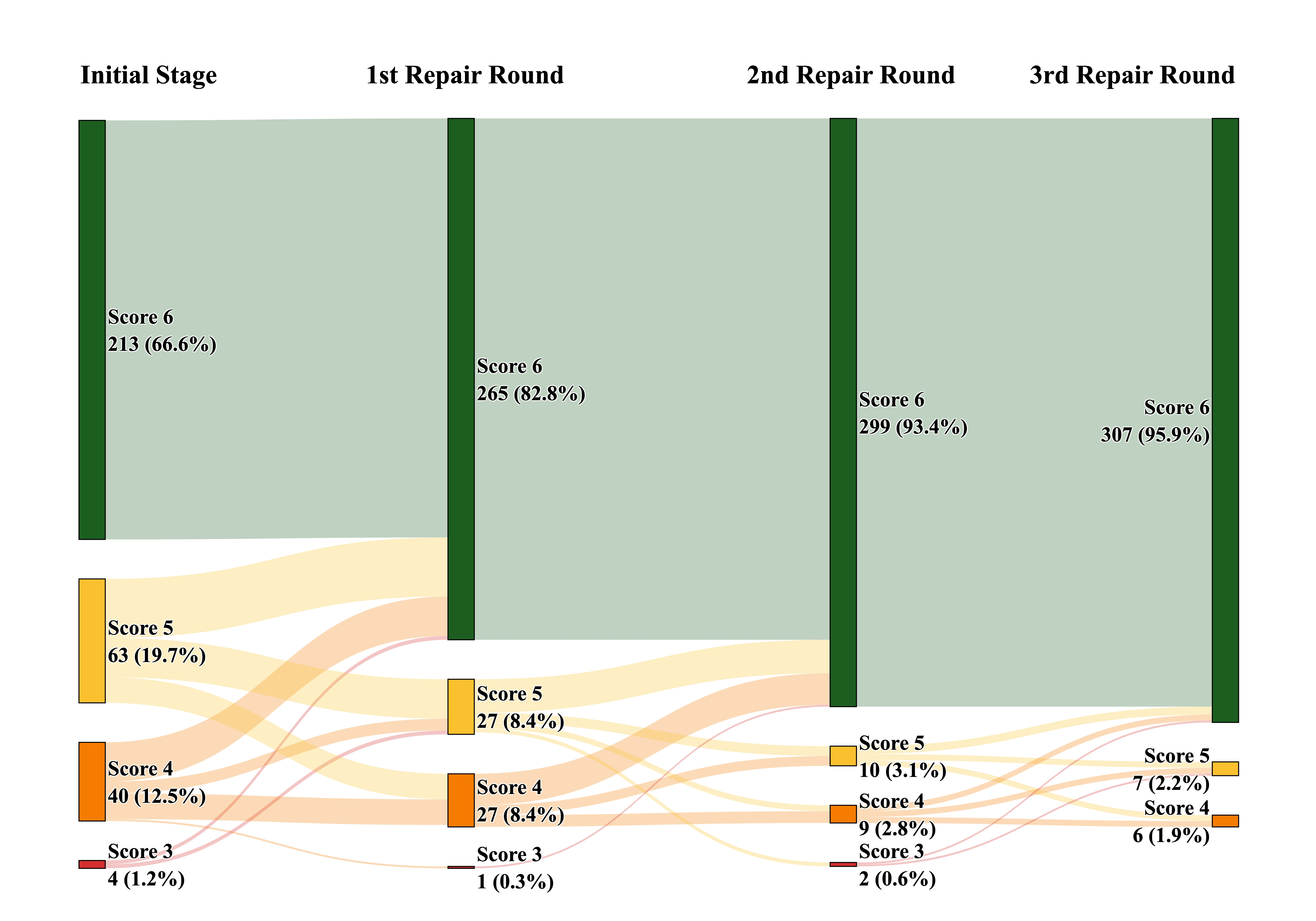}
    \caption{Score transitions across iterative repair rounds.}
    \label{fig:sangji}
\end{figure}

\textbf{Analysis of Generate-Verify-Repair Agent Loop.} 
As shown in the Sankey diagram (Figure~\ref{fig:sangji}), we track score transitions from initial generation to three repair rounds. The accept ratio increases from 66.6\% to 82.8\%, 93.4\%, and 95.9\% across iterations. Although gains gradually saturate and a few samples degrade due to code disruption, the overall trend confirms the effectiveness of fine-grained visual feedback and targeted repair.

\begin{figure*}[!htbp]
    \centering
    
    \subfigure[Qualitative visualization of the Generate-Verify-Repair loop.]{
        \label{fig:case1}
        \includegraphics[width=0.94\linewidth]{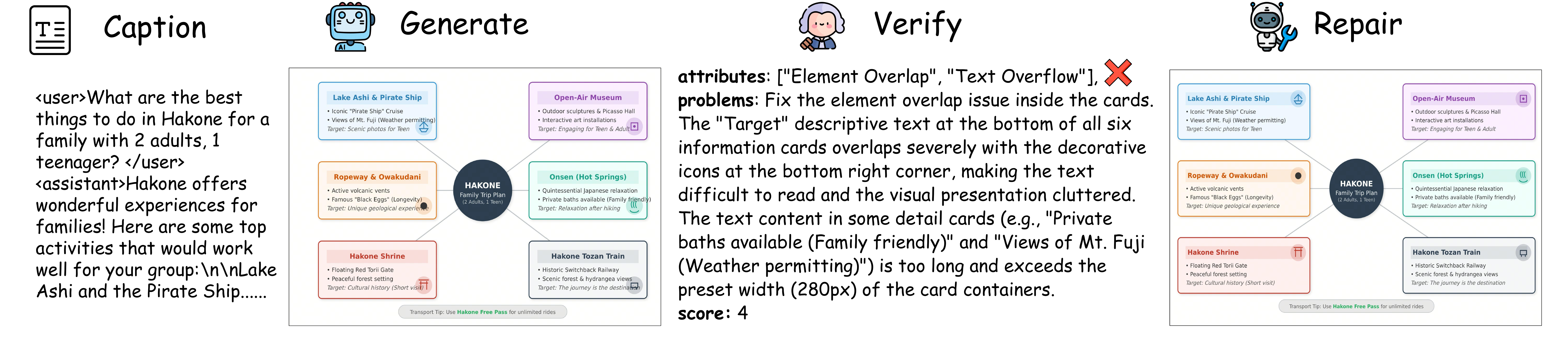}
    }

    \subfigure[Qualitative comparison with SOTA methods.]{
        \label{fig:case2}
        \includegraphics[width=0.94\linewidth]{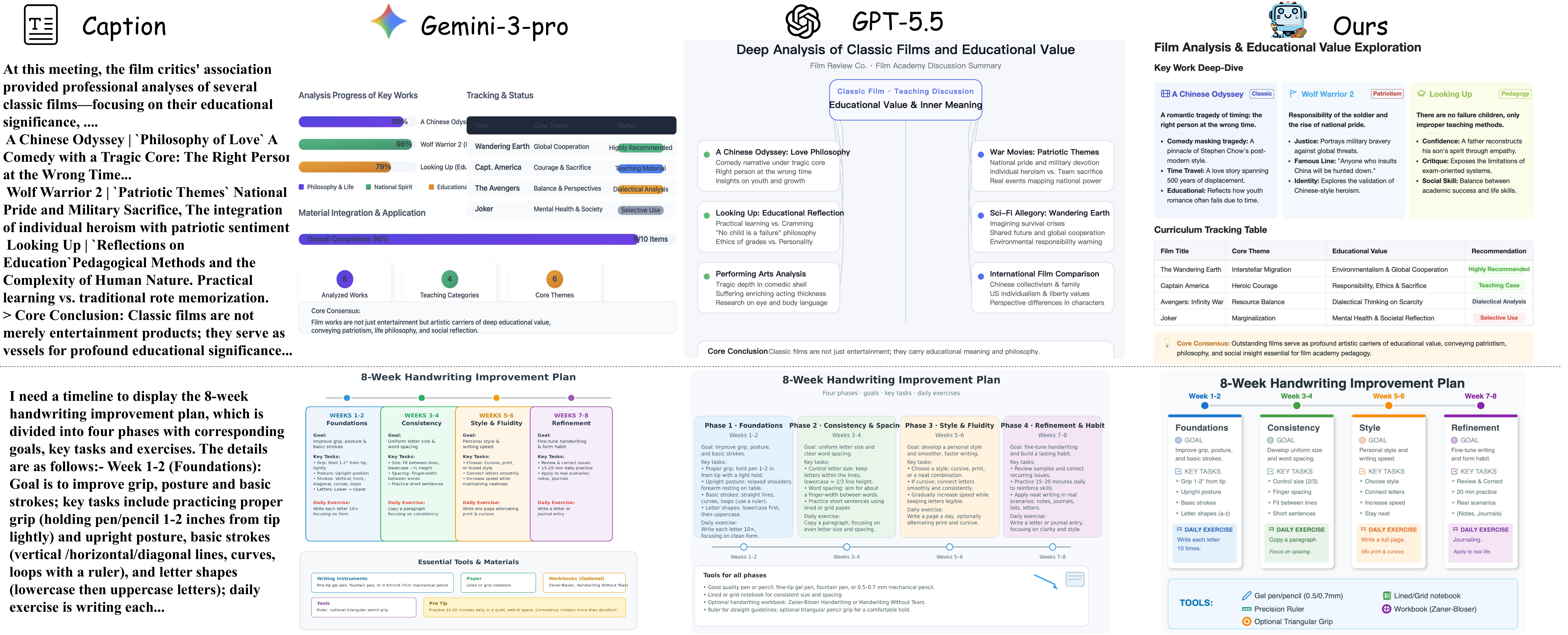}
    }

    \caption{Qualitative performance of GVR-Coder. (Due to space constraints, please refer to the Supplementary Material for high-resolution versions with legible details).} 
    \label{fig:case_study}
\end{figure*}


\subsection{Qualitative Analysis}

\subsubsection{Case Study} 
As shown in Figure \ref{fig:case1}, our iterative agent loop rectifies subtle defects like element overlap, achieving pixel-level precision that single-pass models lack. 
Moreover, qualitative comparisons (Figure \ref{fig:case2}) show our method outperforms closed-source LLMs by avoiding alignment failures and demonstrating superior aesthetic appeal. Notably, via the complexity reward, it prevents over-simplification to ensure robust logic visualization.



\begin{table}[htbp]
\centering
\caption{Human Evaluation. Samples are scored up to a maximum of ten based on criteria detailed in the Supplementary Material. The final reported values are
the average scores.}
\resizebox{\columnwidth}{!}{
\begin{tabular}{l|c|c|c|c|c}
\toprule
\textbf{Metric/Method} & \textbf{Deepseek-V3.2} & \textbf{GPT-5.1} & \textbf{GPT-5.5} & \textbf{Gemini-3-pro} & \textbf{GVR-Coder} \\
\midrule
Semantic Match & 7.77 & 8.78 & 8.26 & 8.89  & \textbf{9.05}\\
Visual Quality &  6.71 & 7.95 & 8.34 & 8.61 & \textbf{8.82} \\
\bottomrule
\end{tabular}
}
\label{tab:human}
\end{table}

\subsubsection{Human Evaluation and User Study} 
To evaluate real-world utility, 10 UI designers scored generated SVGs from 1 to 10 in terms of semantic match and visual quality. Each image was independently evaluated by two designers, with a senior designer resolving discrepancies. As shown in Table~\ref{tab:human}, GVR-Coder achieves the highest average scores in both Semantic Match (9.05) and Visual Quality (8.82), indicating that our method better balances factual consistency and visual aesthetics. 
Conversely, existing LLMs show a 'seesaw effect': Gemini-3 prioritizes semantic logic over aesthetic layouts, 
while GPT-5.5 tends to sacrifice semantic accuracy for better visual appeal.

Beyond expert scoring, we further conduct a blind user study to examine whether the generated diagrams reduce cognitive load in practice. Specifically, 20 participants with diverse ages rank diagrams generated by GVR-Coder, Gemini-3-Pro, GLM-5, and GPT-5.1 on 50 high-cognitive-load texts according to clarity and difficulty reduction. GVR-Coder obtains the highest average ranking score (3.8 vs. 2.8/2.0/1.6), suggesting that its improvements translate into practical readability gains beyond visual quality.

\section{Conclusions}
Generating high-quality, editable diagrams from long-form meeting minutes and document texts holds great promise for office scenarios, while simultaneously posing new challenges to existing SVG generation techniques. To advance this line of research, we introduce DocMeetSVG, a benchmark comprising 100,000 high-quality “office text–svg” pairs that cover diverse document genres and meeting styles.
Building upon this, we further propose GVR-Coder, an SVG agent framework grounded in visual-feedback. We first design a knowledge-enhanced curriculum learning strategy, where structural constraints and hierarchical data jointly guide the model to progressively acquire stable generation capability. We then introduce dual rendering reward signals and integrate them with GRPO-based reinforcement learning to improve completeness and visual appeal. Finally, we develop an agent loop that leverages fine-grained visual feedback and targeted  suggestions to iteratively refine outputs for practical deployment. Through both automatic evaluation and human assessment, we demonstrate that GVR-Coder produces diagrams that meet office standards, 
 achieving both 
 aesthetic layout quality and semantic fidelity.

\begin{acks}
The research was partially supported by the China National Natural Science Foundation with no. 92567301 and 62132018, and Anhui Provincial Science and Technology Innovation Program 202523o09050019 and 202523o09050015.
\end{acks}

\bibliographystyle{ACM-Reference-Format}
\bibliography{ref}

\clearpage

\appendix

\section*{Appendix}

\section{Overview}

In this supplementary material, we provide additional details and evaluations, including:
\begin{itemize}[leftmargin=*]
  \item Data Collection Pipeline.
  \item Detailed Prompt Templates.
  \item Human Evaluation.
  \item Additional Comparisons and Results.
  \item Evaluation Protocols.
  \item Rationale for Adopting SVG as the Target Representation.
  \item More Qualitative Results.
\end{itemize}

\section{Data Collection Pipeline}
To bridge the complexity, logic, and structural gaps in existing SVG datasets for professional scenarios, we introduce a large-scale, structured SVG dataset specifically designed for meeting minutes and office documents. This dataset comprises 100,000 high-quality "office text-SVG diagram" pairs, comprehensively covering diverse Text-to-SVG tasks derived from real-world meeting minutes, natural dialogues, and professional documents.

\textbf{Data Collection.} To ensure authenticity and diversity, text captions are extracted from real-world sources, including meeting records, professional office documents, and the Arena conversation dataset. This multi-source strategy guarantees that the instructions reflect actual requirements in professional collaboration and documentation contexts.


\textbf{Data Inference and Quality Grading.}
To maintain high-fidelity SVG code, we implement an automated yet quality-controlled data construction pipeline. Since different types of diagrams have different structural requirements, we adopt type-specific construction pipelines for different data categories, rather than relying on a single unified prompt template. For each caption, Gemini-3-Pro first generates two independent SVG candidates. A Judge Model then evaluates each candidate through multi-dimensional criteria, including visual correctness, layout clarity, topology validity, and semantic consistency with the input caption.

Based on the scoring results, we stratify the data into three difficulty levels:
\begin{itemize}
    \item \textbf{Simple:} Both samples achieve perfect scores, indicating high generation consistency and relatively stable layout requirements.
    \item \textbf{Medium:} Only one sample reaches the perfect standard, while the other contains minor visual or structural flaws.
    \item \textbf{Hard:} Both samples exhibit flaws, usually corresponding to inputs that require more complex logical reasoning, dense information organization, or sophisticated layout control.
\end{itemize}

\textbf{Hard Instance Repair.}
Identified hard instances are further processed by a specialized Repair Model. Given the imperfect SVG and the feedback from the Judge Model, the Repair Model iteratively revises the SVG code until it satisfies the predefined perfect-output standard. Only samples that pass the final quality check are retained in the dataset.

\textbf{Bias Reduction and Human Verification.}
To reduce construction bias and avoid redundant samples, we conduct deduplication after data generation and repair. In addition, we perform stratified human verification across difficulty levels: 30\% of the samples in each difficulty group are manually audited to check visual quality, semantic faithfulness, and layout correctness. This ensures that the dataset quality is not solely determined by automatic model judgments. For evaluation, the 320 test cases are collected from real documents and meeting records outside the training set, and all test samples are fully manually validated before being used for benchmarking.

\textbf{Data Distribution.}The final dataset comprises 100,000 high-standard samples. The data spans six primary categories: \textit{Explanatory Vis}, \textit{Document Authoring}, \textit{Meeting Summary}, \textit{Data Vis}, \textit{Others}, and \textit{Icons}. Notably, \textit{Explanatory Vis} constitutes the largest portion with 59,137 samples. Its granular sub-categories include Flowcharts (17.3\%) and Architectural Diagrams (11.6\%), which meet the demand for generating complex, document-level logical illustrations.
Experiments demonstrate that this dataset effectively enhances model performance in visual semantic understanding and SVG code generation tasks, providing a solid foundation for related research.

\section{Detailed Prompt Templates}

\begin{figure}[htbp]
    \centering
    \begin{tcolorbox}[
        colframe=green!40!black,
        colback=green!5!white,
        boxrule=1pt,
        arc=4pt,
        outer arc=4pt,
        boxsep=5pt,
        left=6pt,
        right=6pt,
        top=6pt,
        bottom=6pt,
        fonttitle=\bfseries,
        title=Prompt: SVG Generator Agent,
        ]
    \small
    You are an AI assistant specialized in generating SVG illustrations.

    \textbf{Requirements:}
    \begin{itemize}[leftmargin=*]
        \item Generate a professional SVG based on the provided user text, scenario (e.g., meeting review or office report), and diagram type (e.g., Flowchart, Mind Map, or Meeting Cover figure).
        \item The style must be professional, clean, and high-contrast.
    \end{itemize}

    \textbf{Constraint Knowledge Checklist (must avoid):}
    \begin{itemize}[leftmargin=*]
        \item Element overlap (nodes, text, or edges intersecting or occluding each other).
        \item Incorrect connections (wrong arrow direction, broken/misaligned links, illogical connections, overly complex paths).
        \item Text overflow (text exceeding container boundaries or improperly wrapped).
        \item Cropped content (improper viewBox or insufficient canvas margins causing clipping).
        \item Severe alignment or style inconsistency (misalignment, uneven spacing, inconsistent styling).
        \item Factual inconsistency (fabricating ungrounded information, omitting key entities, or logical contradictions with the source text).
    \end{itemize}

    Ensure the generated SVG code satisfies all constraints and contains none of the above issues.
        
        \textbf{Output Format:}
    \begin{itemize}[leftmargin=*]
        \item The output must be a complete SVG code block.
        \item The code must be concise, well-structured, and visually clean.
        \item The SVG must be written in valid XML format.
    \end{itemize}

\begin{verbatim}
<svg xmlns="http://www.w3.org/2000/svg" viewBox="0 0 X Y">
...
</svg>
\end{verbatim}
    \end{tcolorbox}
    \caption{Prompt used for structured SVG generation.}
    \label{fig:generator_prompt}
\end{figure}

\begin{figure}[htbp]
    \centering
    \begin{tcolorbox}[
        colframe=red!40!black,
        colback=red!5!white,
        boxrule=1pt,
        arc=4pt,
        outer arc=4pt,
        boxsep=5pt,
        left=6pt,
        right=6pt,
        top=6pt,
        bottom=6pt,
        fonttitle=\bfseries,
        title=Prompt: SVG Verify Agent,
        ]
    \small

    \textbf{Role.}  
    You are an experienced design consultant specializing in SVG standards, frontend visualization engineering, and advanced visual aesthetics.

    \textbf{Objective.}
    \begin{itemize}[leftmargin=*, itemsep=2pt]
        \item Evaluate the rendered SVG result based on the caption, SVG code, and rendered image.
        \item Identify severe issues and attribute them to predefined categories.
        \item Determine whether the SVG requires repair.
        \item Provide concise and actionable improvement suggestions.
    \end{itemize}

    \textbf{Predefined Issue Categories.}
    \begin{itemize}[leftmargin=*, itemsep=2pt]
        \item Element overlap
        \item Severe alignment/style issues
        \item Incorrect connections
        \item Unreasonable layout (imbalanced or chaotic structure)
        \item Text overflow
        \item Cropped or incomplete content
    \end{itemize}

    \textbf{Evaluation Principles.}
    \begin{itemize}[leftmargin=*, itemsep=2pt]
        \item Focus only on severe issues that affect readability or visual integrity.
        \item Minor pixel-level misalignment or stylistic variations do not require repair.
    \end{itemize}

    \textbf{Output Format.}  
    Output must strictly follow the structured XML format:

    \textit{If repair is required:}

\begin{verbatim}
<attribute>
[Issue Name]
</attribute>
<problems>
[Concise analysis and targeted suggestions]
</problems>
<is_repaired>Yes</is_repaired>
\end{verbatim}

    \textit{If no repair is required:}

\begin{verbatim}
<is_repaired>No</is_repaired>
\end{verbatim}

    \end{tcolorbox}
    \caption{Prompt used for SVG aesthetic and standardization evaluation.}
    \label{fig:critic_prompt}
\end{figure}

\begin{figure}[htbp]
    \centering
    \begin{tcolorbox}[
        colframe=purple!40!black,
        colback=purple!5!white,
        boxrule=1pt,
        arc=4pt,
        outer arc=4pt,
        boxsep=5pt,
        left=6pt,
        right=6pt,
        top=6pt,
        bottom=6pt,
        fonttitle=\bfseries,
        title=Prompt: Factual Consistency Evaluator,
        ]
    \small

    \textbf{Role.}  
    You are a professional content audit expert specializing in evaluating the factual consistency between generated diagrams and their source text (Caption).

    \textbf{Objective.}  
    Based on the provided Caption (Ground Truth) and the rendered image, determine if the image is factually accurate across three dimensions. 
    Assign a binary factual score: \textbf{0 or 1}.

    \textbf{Evaluation Dimensions.}
    The image must \textbf{simultaneously satisfy} all three dimensions to receive a score of 1:
    \begin{itemize}[leftmargin=*, itemsep=2pt]
        \item \textbf{1. Zero Hallucination (Strict Anchoring):} Every text node must have a direct source in the caption. No inferred intermediate steps or principles. \textit{Whitelist:} English translations, Subgraph containers, "End" nodes, and accurate summary titles.
        \item \textbf{2. No Omission:} Core stakeholders, key steps, and explicit end states must be present. (Exception: Data from logically different dimensions/scales are not considered missing).
        \item \textbf{3. Logical Consistency:} Hierarchical branching must use consistent classification dimensions. The diagram type (Flowchart, Sequence, etc.) and connection relationships must strictly align with the caption.
    \end{itemize}

    \textbf{Scoring Criteria.}
    \begin{itemize}[leftmargin=*, itemsep=2pt]
        \item \textbf{Score 1:} The image faithfully meets all three dimensions without any hallucination, omission, or logical error.
        \item \textbf{Score 0:} The image contains any hallucination (outside the whitelist), missing core content, or hierarchical/relational inconsistencies.
    \end{itemize}

    The score must be strictly binary (0 or 1).

    \textbf{Output Format.}  
    First provide a detailed analysis of the three dimensions within \texttt{<reason></reason>}, then provide the final score within \texttt{<score></score>}.

\begin{verbatim}
<reason>
- Hallucination Analysis: ...
- Omission Analysis: ...
- Logic & Hierarchy Analysis: ...
</reason>
<score>0 or 1</score>
\end{verbatim}

    No additional commentary is allowed.

    \end{tcolorbox}
    \caption{Prompt for factual consistency evaluation.}
    \label{fig:factual_prompt}
\end{figure}


\begin{figure}[htbp]
    \centering
    \begin{tcolorbox}[
        colframe=blue!40!black,
        colback=blue!5!white,
        boxrule=1pt,
        arc=4pt,
        outer arc=4pt,
        boxsep=5pt,
        left=6pt,
        right=6pt,
        top=6pt,
        bottom=6pt,
        fonttitle=\bfseries,
        title=Prompt: SVG Repair Agent,
        ]
    \small

    \textbf{Role.}  
    You are a rigorous SVG repair engineer and code optimization expert.

    \textbf{Objective.}  
    Given the original caption, the initial svg code, the Verify Agent’s attribute tags, and the detailed problems analysis, strictly follow the Verify Agent’s feedback and precisely fix all identified issues while preserving correct semantics and structure.

    \textbf{Core Tasks.}
    \begin{itemize}[leftmargin=*, itemsep=2pt]
        \item Convert each reported issue into concrete code-level modifications (e.g., adjusting coordinates, sizes, paths, alignment, text layout).
        \item Recalculate layout for alignment or spacing problems.
        \item Ensure responsive design by optimizing the viewBox and removing fixed width/height constraints.
        \item Eliminate overlap, overflow, clipping, and connection errors completely.
    \end{itemize}

    \textbf{Quality Requirements.}
    \begin{itemize}[leftmargin=*, itemsep=2pt]
        \item Maintain semantic accuracy with respect to the caption.
        \item Ensure visual clarity, modern flat style, and aesthetic consistency.
        \item Produce clean, well-structured, fully renderable SVG code.
    \end{itemize}

    \textbf{Output Constraint.}  
    Return only the repaired SVG wrapped in:

\begin{verbatim}
<repaired_svg>
<svg ...>
...
</svg>
</repaired_svg>
\end{verbatim}

    No additional explanation or text is allowed.

    \end{tcolorbox}
    \caption{Prompt used for SVG repair and refinement.}
    \label{fig:repair_prompt}
\end{figure}

The SVG Generator prompt defines the core generation policy of our system, as shown in Figure~\ref{fig:generator_prompt} It specifies structural constraints, visual style requirements, and a constraint knowledge checklist to prevent common layout and rendering errors. By explicitly encoding design rules and output formatting requirements, this prompt guides the model to produce clean, well-structured, and fully renderable SVG code under standardized generation constraints.

The SVG Verify prompt functions as a structured evaluation agent (Figure \ref{fig:critic_prompt}). It analyzes the rendered SVG result with respect to predefined issue categories, including overlap, misalignment, layout disorder, and structural inconsistencies. By focusing exclusively on severe issues that compromise readability or visual integrity, the verify agent ensures objective feedback for downstream refinement.

As shown in Figure \ref{fig:factual_prompt}, the Factual Consistency prompt evaluates the semantic alignment between the input caption,  and the rendered SVG image across three dimensions: Zero Hallucination (preventing unmentioned information), No Omission (ensuring core elements are present), and Logical Consistency (verifying hierarchical and structural accuracy). It performs element-level matching and assigns a binary factual score, ensuring the generated SVGs remain strictly faithful to the intended description.

The SVG Repair prompt implements a targeted correction mechanism (Figure \ref{fig:repair_prompt}). Guided by the Verify Agent’s feedback, it converts abstract issue descriptions into concrete code-level modifications, recalculates layout when necessary, and eliminates structural and visual defects. The repair agent strictly outputs corrected SVG code, forming a closed-loop refinement pipeline.

\section{Human Evaluation}

As VLM-based evaluation cannot fully capture the subtleties of human aesthetics, we invited 10 UI designers to perform assessment. 
To ensure objectivity, each image was independently rated by two evaluators. If significant scoring discrepancies occurred, a senior UI designer intervened to make the final determination. 
The assessment followed a
strict double-blind testing procedure, with each evaluator independently completing professional judgments on the following two
dimensions:
\begin{itemize}[leftmargin=*]
    \item \textbf{Semantic Match:} Measures the semantic alignment between the generated SVG and the input text across three dimensions: Zero Hallucination, No Omission, and Logical Consistency. A score of 1 is awarded only if the SVG satisfies all three criteria, whereas a score of 0 is assigned if it contains hallucinations, missing content, or structural errors. To ensure comparability, the final average score is normalized to a 10-point scale.

    \item \textbf{Visual Quality:} Evaluates structural and aesthetic integrity by identifying specific visual defects, including element overlap, connection issues, text overflow, color coordination, and content occlusion. A deductive scoring mechanism is employed based on the presence and severity of these errors, where any severe defect precludes a high score. The final evaluation is mapped to a discrete scale of 10, 8, 6, 4, or 2 points.
\end{itemize}

\begin{figure*}[htbp]
    \centering
    \includegraphics[width=1.0\textwidth]{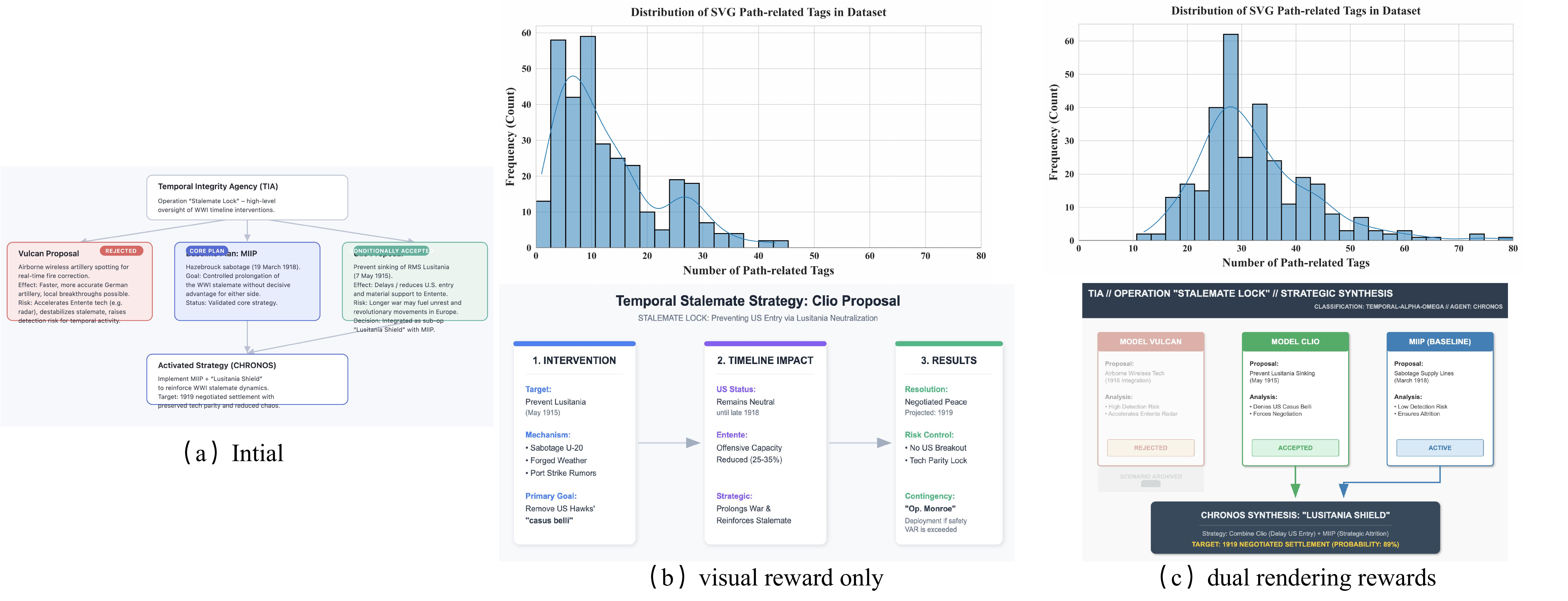}
    \caption{Effectiveness of dual rendering rewards.}
    \label{fig:rewards}
\end{figure*}

\begin{figure}[h]
\centering  
\subfigure[SVG visual reward]{\label{fig:svgscore}
{\includegraphics[width=3.96cm,totalheight=3.2cm]{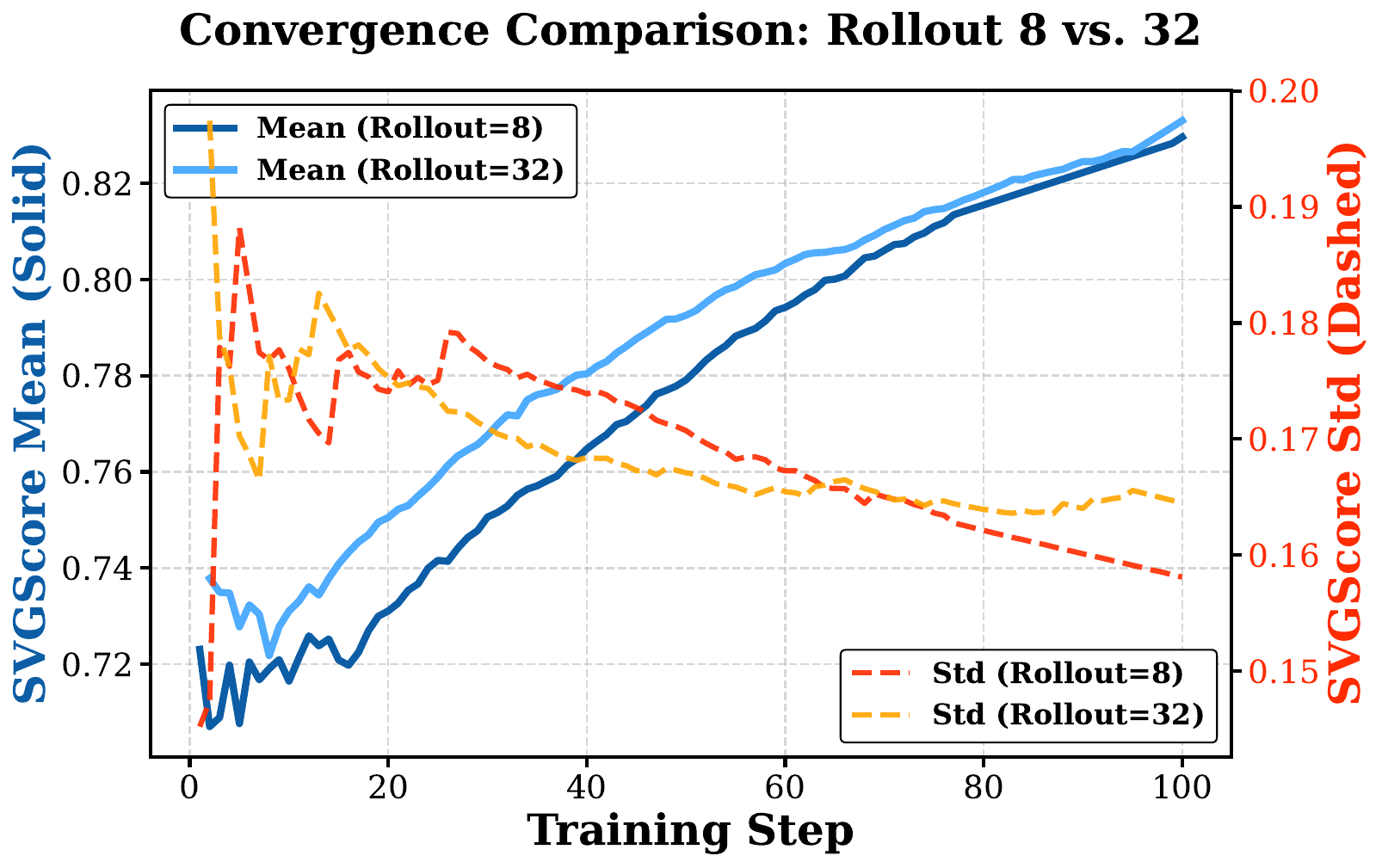}}}
\quad
\subfigure[SVG complexity reward]{\label{fig:svgcom}
{\includegraphics[width=3.97cm,totalheight=3.2cm]{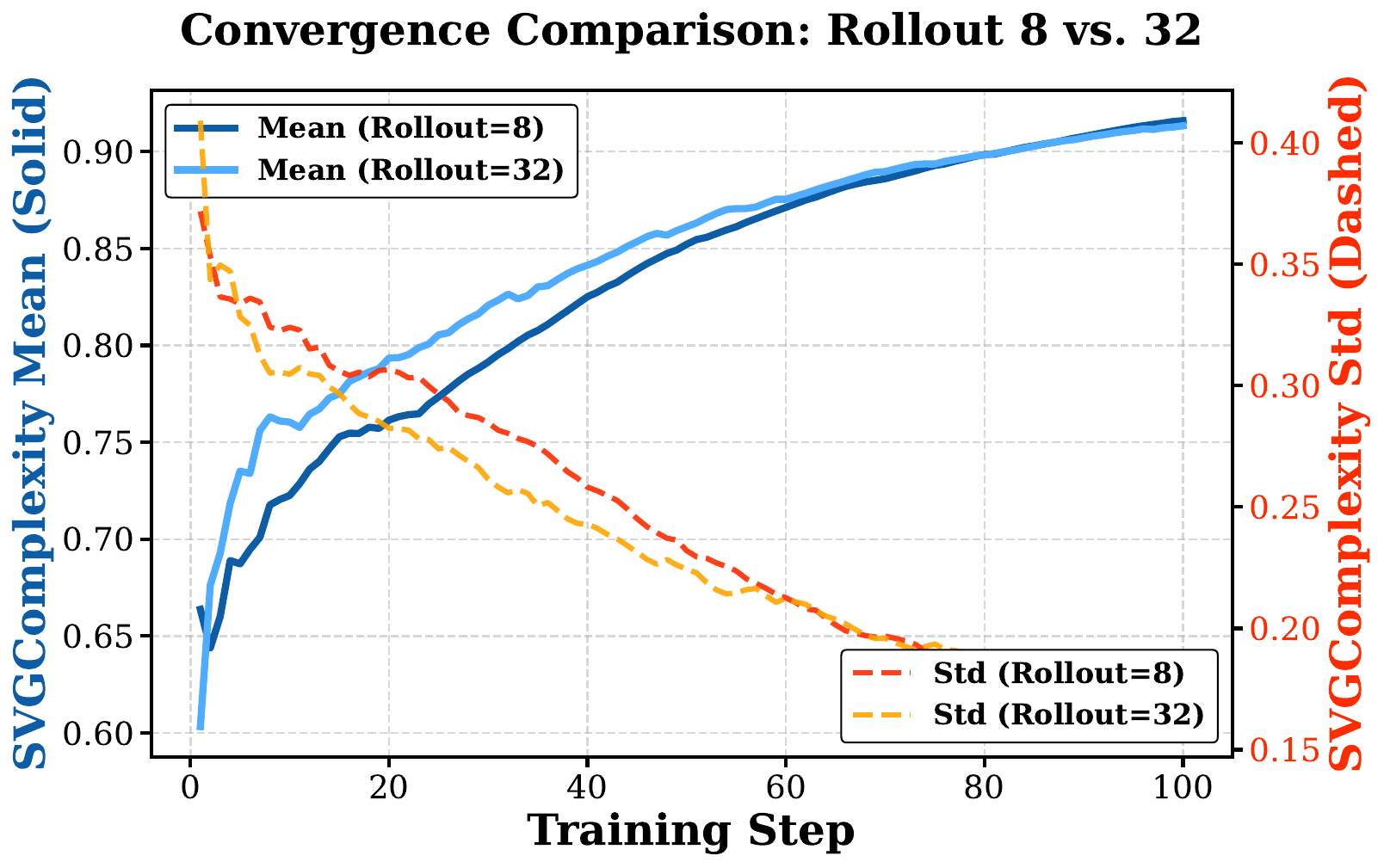}}}
\quad
\subfigure[Completion Length of RLDRF]{\label{fig:len}
{\includegraphics[width=3.8cm,totalheight=3.2cm]{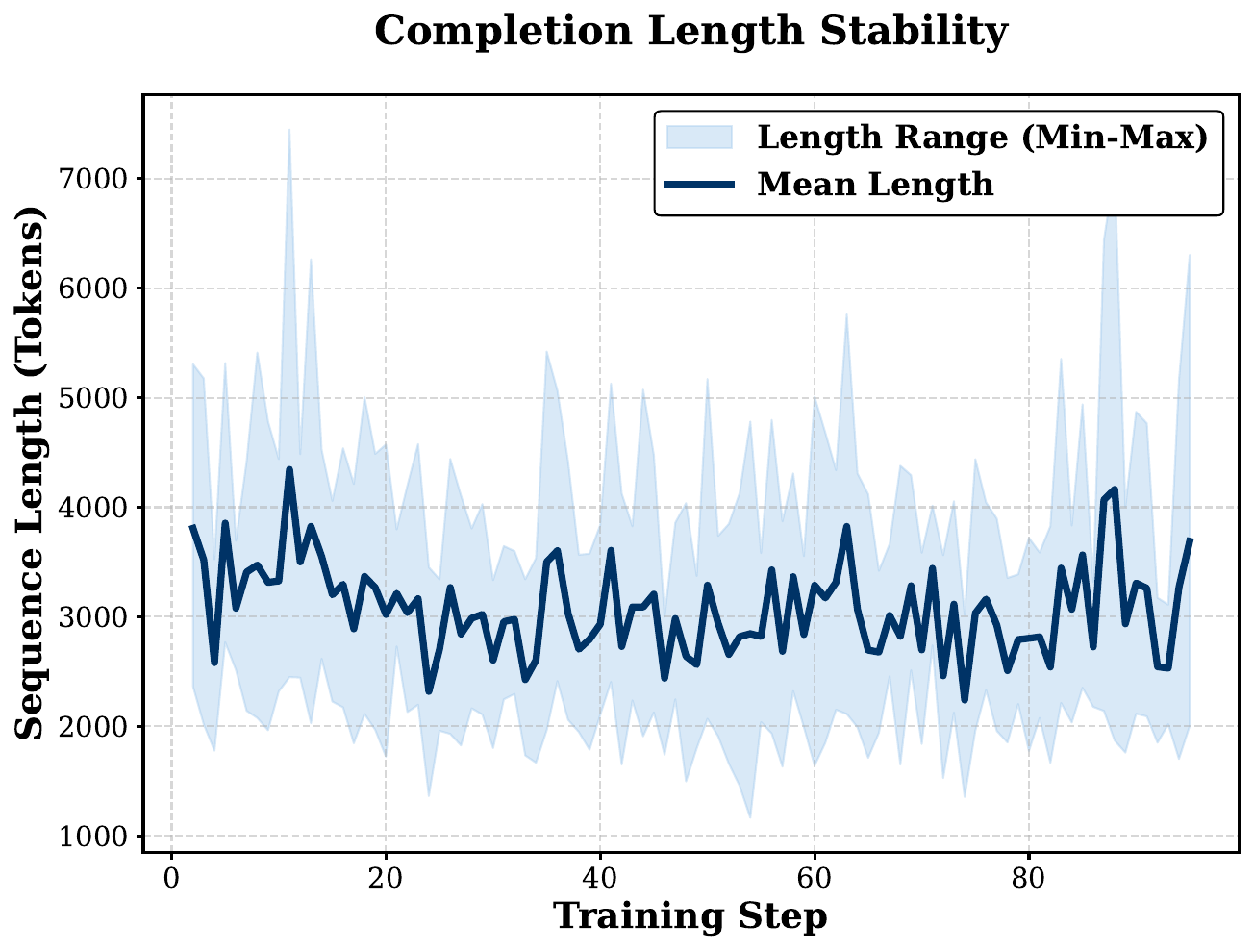}}}
\quad
\subfigure[Performance evolution of RLDRF]{\label{fig:dpo}
{\includegraphics[width=3.8cm,totalheight=3.2cm]{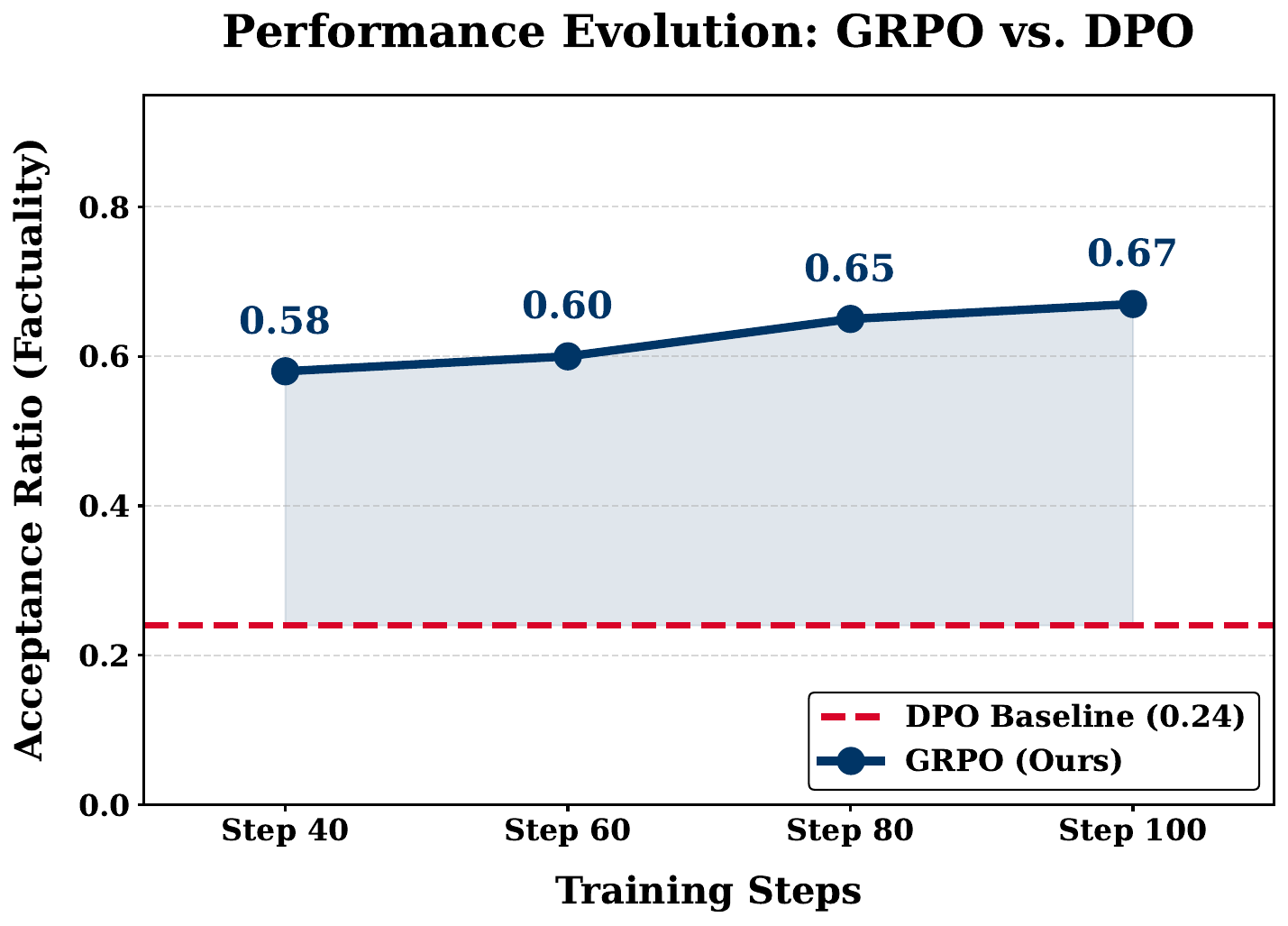}}}
\caption{\small Analysis of RLDRF Training Dynamics on n320 datasets.} \label{fig-tal}
\end{figure}

    
    

\section{Additional Comparisons and Results}
We provide additional quantitative results on the DocMeetSVG-100K dataset.

\textbf{Analysis of Dual Rendering Rewards.} To further validate the necessity of our dual-reward mechanism, we qualitatively and quantitatively compare the generation results under different reward configurations (Figure \ref{fig:rewards}). While the initial model (a) suffers from visually unappealing text overflow and rudimentary layouts, optimizing solely with a visual reward (b) inadvertently encourages the model to exploit shortcuts. This phenomenon, which we term "over-simplification," is clearly reflected in the skewed distribution of SVG path-related tags (Figure \ref{fig:rewards} (b), top), where the model heavily favors overly simplistic structures to easily satisfy basic aesthetic constraints. 

In contrast, the integration of dual rendering rewards acts as a rigorous structural constraint that effectively balances graphical complexity and aesthetic layout. By penalizing such shortcuts, the path tag distribution smoothly shifts toward higher complexity (Figure \ref{fig:rewards} (c), top), guiding the model to generate structurally rich diagrams. This stable optimization process is further corroborated by our training metrics. As shown in Figures \ref{fig:svgscore} and \ref{fig:svgcom}, both visual and complexity rewards exhibit steady growth, surpassing 0.82 and 0.90 respectively. Notably, a larger rollout size (Rollout=32) yields more stable convergence with a faster reduction in variance. Crucially, Figure \ref{fig:len} demonstrates that completion lengths remain stable (3,000–4,000 tokens) throughout this process, firmly indicating that the model optimizes spatial layouts without resorting to degenerate shortcuts to inflate visual scores. Finally, the robustness of our approach is evident in the accept ratio, where GVR-Coder improves steadily to 0.67 (Figure \ref{fig:dpo}), outperforming the DPO baseline (0.24). This gap stems from DPO's offline nature, which evaluates sequences holistically and fails to assign precise credit to the specific local tokens responsible for visual errors.

\begin{figure}[htbp]
    \centering
\includegraphics[width=0.5\textwidth]{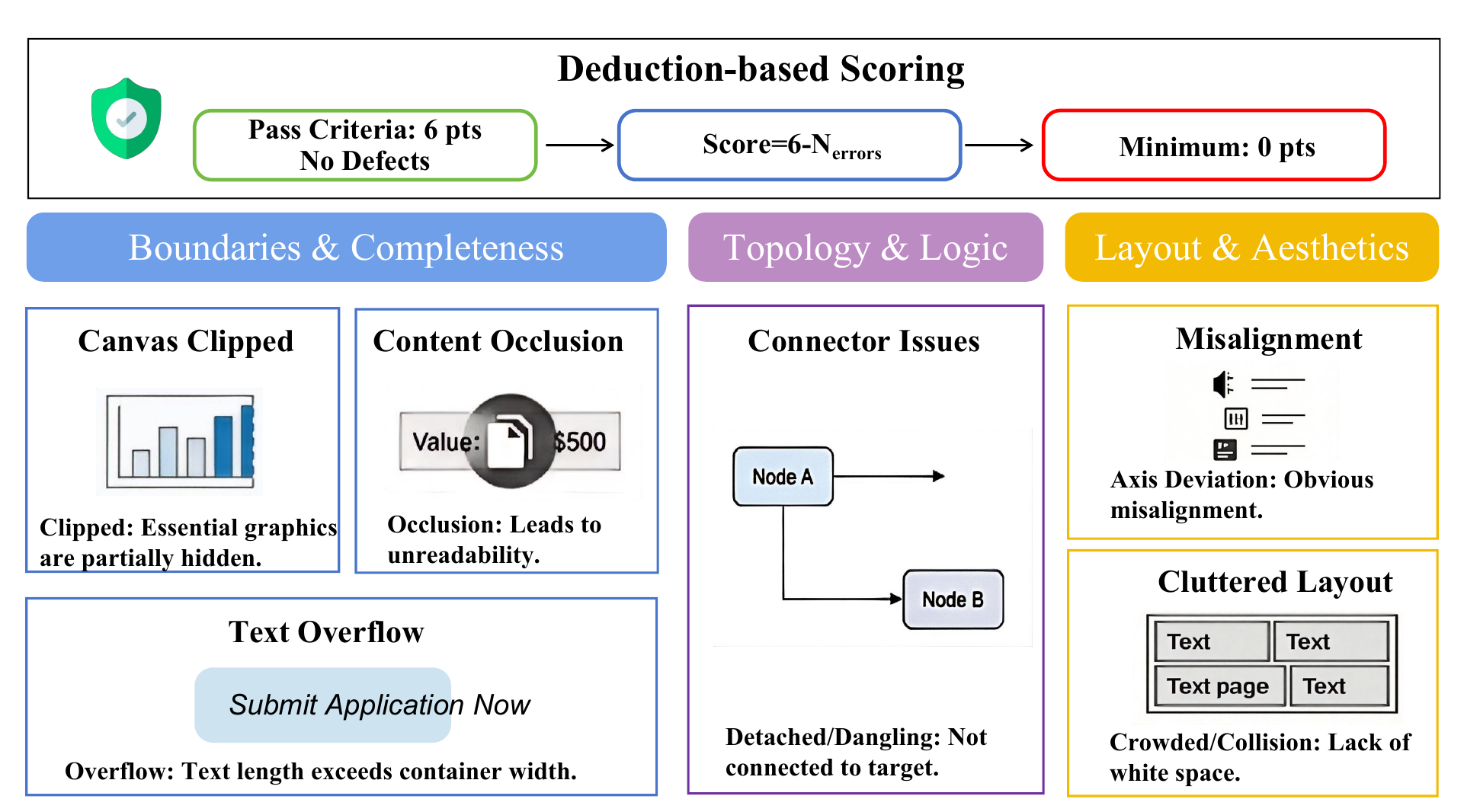}
    \caption{Quantitative scoring mechanism and multi-dimensional defect taxonomy.}
    \label{fig:error_taxonomy}
\end{figure}

\section{Evaluation Protocols}

\textbf{Aesthetic and Normative Scoring}:
Standard Text-to-Image metrics such as FID and CLIPScore often correlate poorly with the logical and topological requirements of document-level diagrams. We therefore adopt Gemini-3-Pro as VLM-as-Judge to provide nuanced visual feedback.

To bridge the gap between automated checking and manual review, we classified errors into three dimensions and six types (Figure \ref{fig:error_taxonomy}). We utilize a deductive scoring mechanism:
$Score = \max(0, 6 - N)$
where $N$ represents the number of detected errors.

\textbf{Factuality Scoring}:
The score is assigned to measure semantic alignment across three core dimensions: Zero Hallucination, No Omission, and Logical Consistency.
1 Point: The SVG satisfies all three criteria, faithfully reflecting the caption without redundant content.
0 Point: The image fails in any dimension. For more detailed prompts, please refer to the supplementary materials.

\section{Rationale for Adopting SVG as the Target Representation}
\label{sec:why_svg}

We adopt Scalable Vector Graphics as our primary diagram representation due to its unique balance of visual expressiveness, precise spatial control, and widespread cross-platform compatibility. 

While declarative graph languages (e.g., Mermaid) are efficient for standard charts, they are heavily constrained by predefined routing algorithms and fixed templates. In contrast, SVG natively supports free-form layouts, nested architectural cards, customized iconography, and complex visual styling such as gradients, enabling fine-grained design far beyond rigid typologies. 

Additionally, compared to web-native formats like HTML/CSS, SVG operates on an explicit, self-contained 2D coordinate system. This mathematical foundation decouples the diagram from browser-specific DOM rendering or external stylesheets. Consequently, SVG guarantees higher spatial freedom and native scalability, allowing the generated assets to be seamlessly inserted, resized, and manually refined in diverse downstream environments, including documents, presentation slides, and vector graphics editors. Ultimately, SVG serves as a practical medium that bridges the gap between programmatic logic generation and aesthetic visual design.

\section{More Qualitative Results}

\begin{figure*}[htbp]
    \centering
    \includegraphics[width=1.0\textwidth]{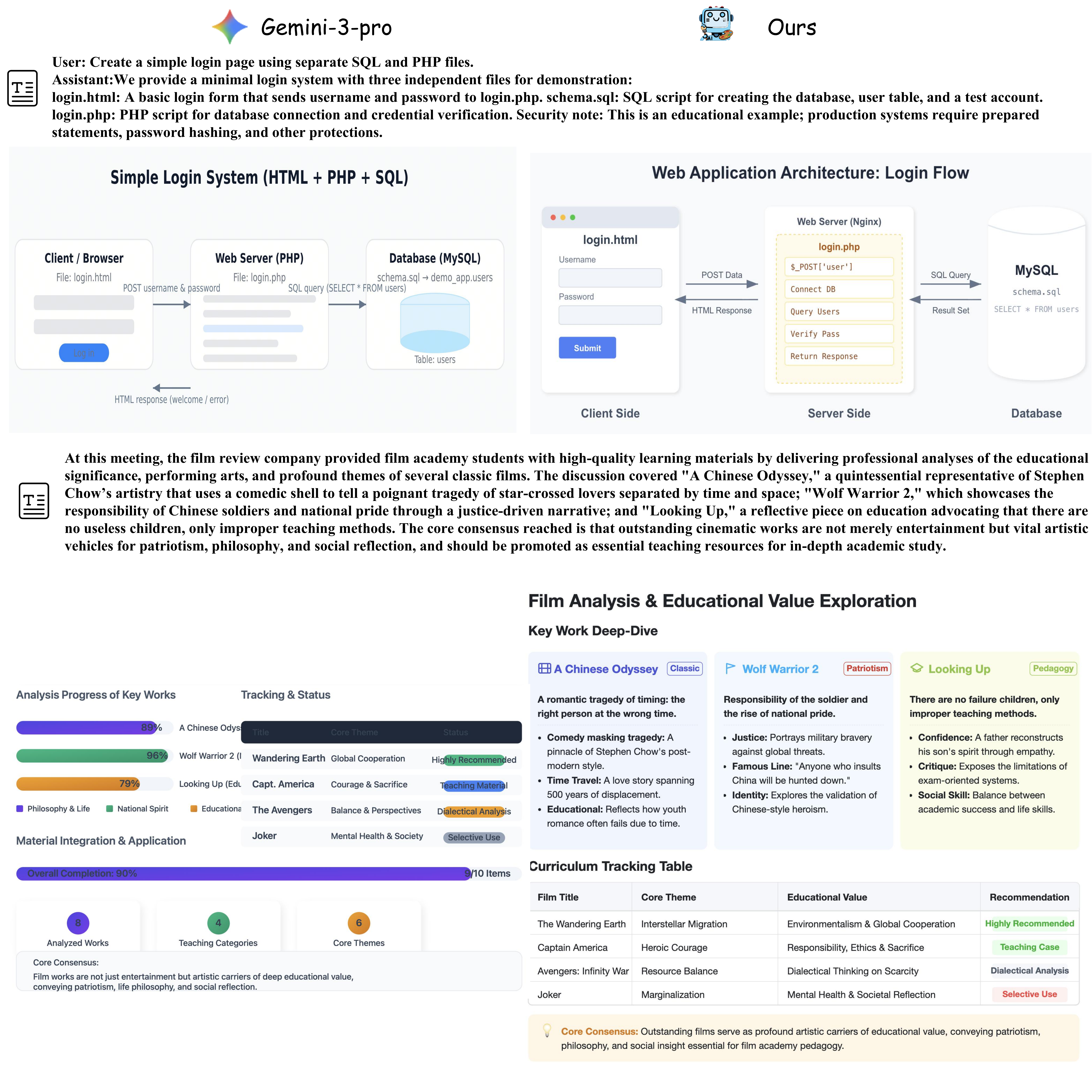}
    \caption{Comparative visualization results between Gemini-3-pro and GVR-Coder}
    \label{fig:vis1}
\end{figure*}

\begin{figure*}[htbp]
    \centering
    \includegraphics[width=1.0\textwidth]{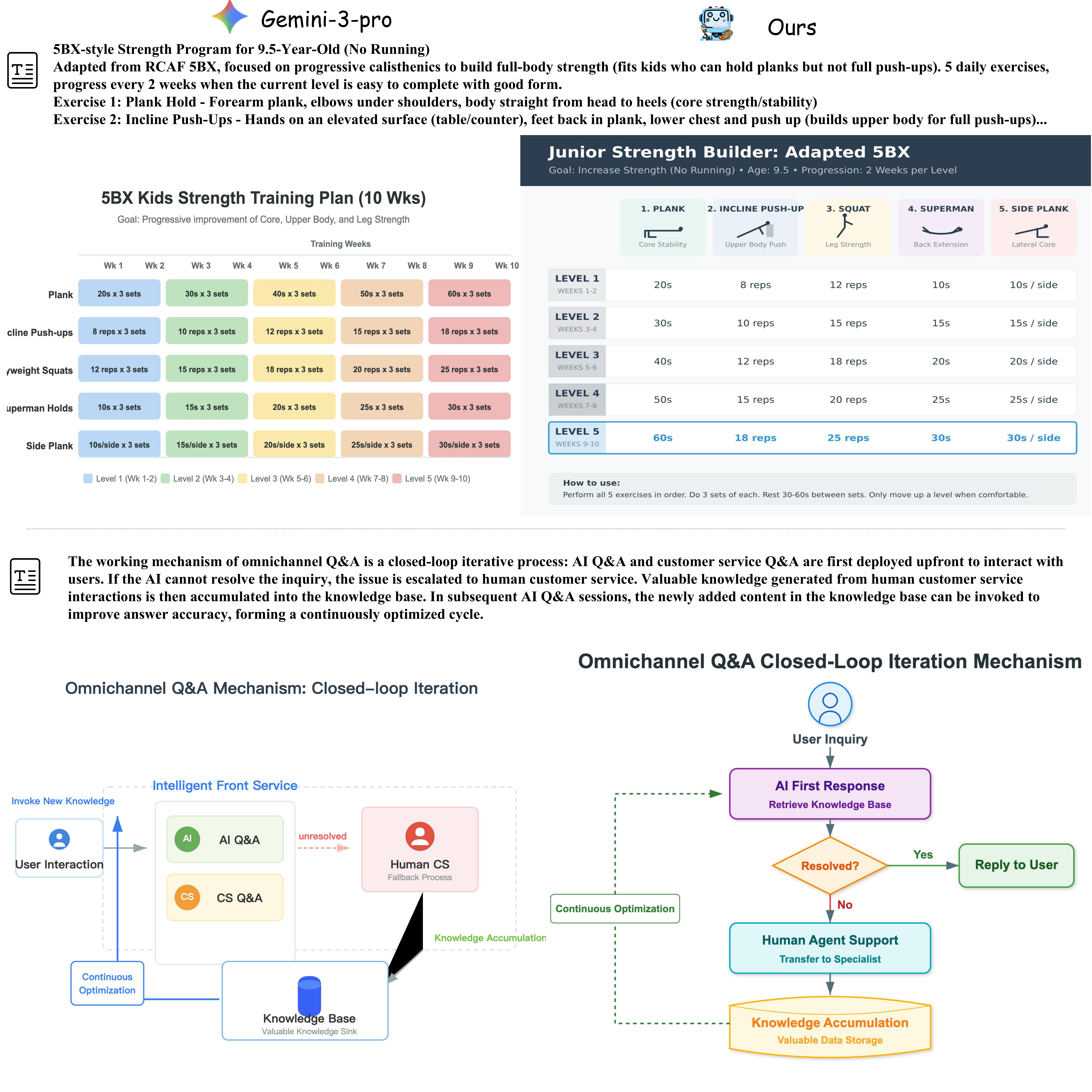}
    \caption{Comparative visualization results between Gemini-3-pro and GVR-Coder}
    \label{fig:vis2}
\end{figure*}

\begin{figure*}[htbp]
    \centering
    \includegraphics[width=1.0\textwidth]{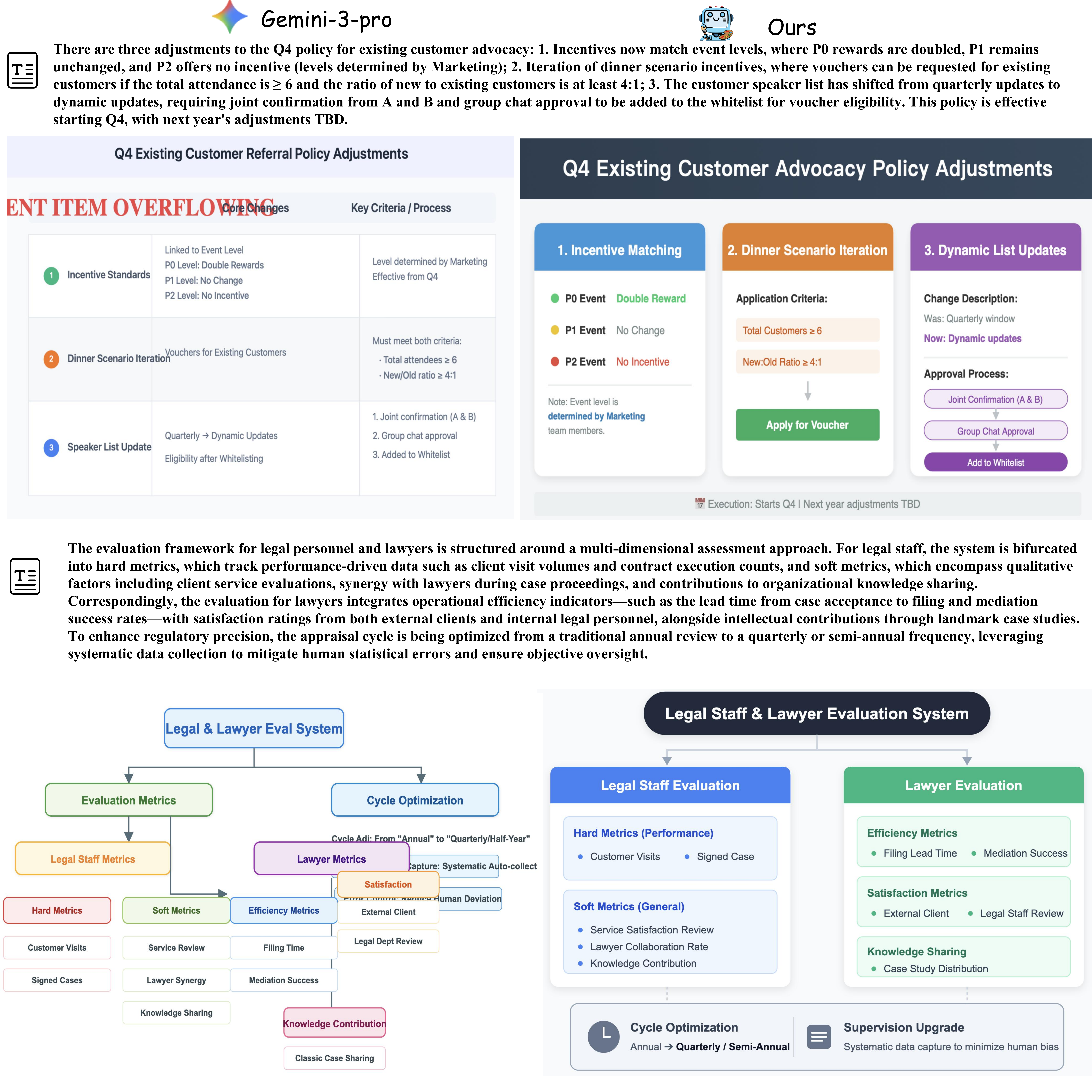}
    \caption{Comparative visualization results between Gemini-3-pro and GVR-Coder}
    \label{fig:vis_3}
\end{figure*}

\begin{figure*}[htbp]
    \centering
    \includegraphics[width=1.0\textwidth]{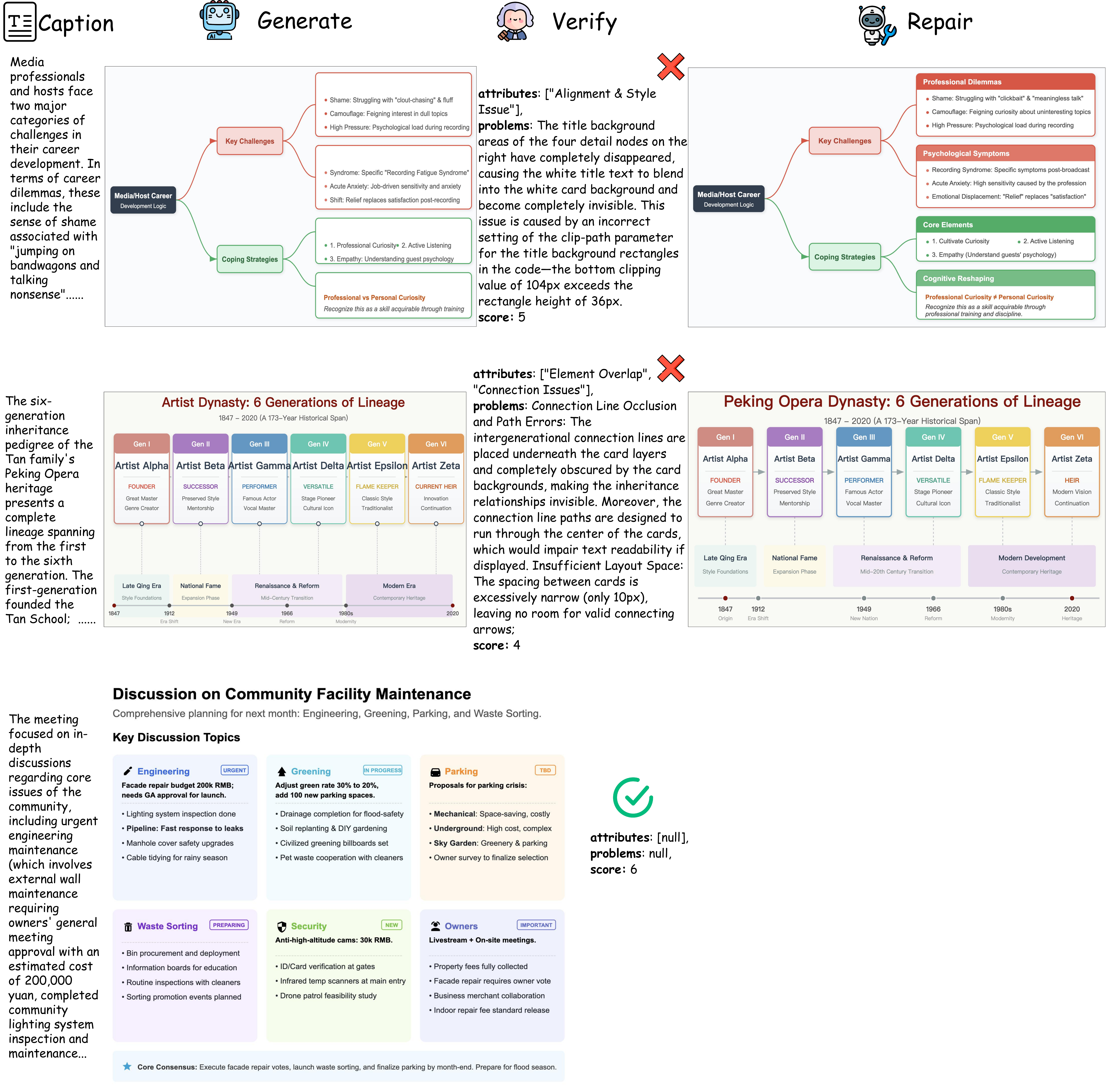}
    \caption{Visualization of the Generate-Verify-Repair Agent Loop (VRL) process}
    \label{fig:vis_gvr}
\end{figure*}

In this section, we provide additional qualitative visualizations to further demonstrate the effectiveness and robustness of our proposed framework. Figure \ref{fig:vis1}, \ref{fig:vis2}, and \ref{fig:vis_3} present an extended visual comparison between our model and state-of-the-art closed-source LLMs (i.e., Gemini-3-pro) across diverse and complex diagram types. The results clearly highlight our model's superiority in maintaining strict layout alignment, hierarchical clarity, and aesthetic design. Furthermore, Figure \ref{fig:vis_gvr} provides detailed, step-by-step examples of the Generate-Verify-Repair agent loop in action. It illustrates how the verifier model accurately identifies specific visual defects—such as missing background elements or overlapping connection lines—and effectively guides the repair model to produce highly polished, deployment-ready SVGs.

\end{document}